\documentclass[runningheads]{llncs}

\usepackage{graphicx}
\usepackage{comment}
\usepackage{amsmath,amssymb}
\usepackage{color}
\usepackage{url}
\usepackage{hyperref}

\usepackage{xspace}

\makeatletter
\DeclareRobustCommand\onedot{\futurelet\@let@token\@onedot}
\def\@onedot{\ifx\@let@token.\else.\null\fi\xspace}

\def\etal{\emph{et al}\onedot}
\makeatother

\usepackage{mathtools}
\usepackage{booktabs}

\newif\ifreview
\reviewfalse

\ifreview
	\usepackage{lineno}

	\linenumbers
\fi

\begin{document}


\def\SubNumber{150}

\def\GCPRTrack{Fast Review Track}

\title{FIFA: Fast Inference Approximation for Action Segmentation}

\ifreview
\else

	\author{Yaser Souri\inst{1} \and
	Yazan Abu Farha\inst{1} \and
	Fabien Despinoy\inst{2} \and
	Gianpiero Francesca\inst{2} \and
	Juergen Gall\inst{1}
	}
	
	\authorrunning{Y. Souri et al.}
	
	\institute{University of Bonn \and Toyota Motor Europe\\
	\email{\{souri, abufarha, gall\}@iai.uni-bonn.de} \\ \email{\{fabien.despinoy, gianpiero.francesca\}@toyota-motor.com}}
\fi

\maketitle              

\begin{abstract}
We introduce FIFA, a fast approximate inference method for action segmentation and alignment. Unlike previous approaches, FIFA does not rely on expensive dynamic programming for inference. Instead, it uses an approximate differentiable energy function that can be minimized using gradient-descent. FIFA is a general approach that can replace exact inference improving its speed by more than 5 times while maintaining its performance. FIFA is an anytime inference algorithm that provides a better speed vs. accuracy trade-off compared to exact inference. We apply FIFA on top of state-of-the-art approaches for weakly supervised action segmentation and alignment as well as fully supervised action segmentation. FIFA achieves state-of-the-art results on most metrics on two action segmentation datasets.

\end{abstract}
\section{Introduction}


Action segmentation is the task of predicting the action label for each frame in the input video. Action segmentation is usually studied in the context of activities performed by a single person, where temporal smoothness of actions are assumed. Fully supervised approaches for action segmentation~\cite{lea2017temporal,MS-TCN,li2020ms,wang2020boundary} already achieve good performance on this task. Most approaches for fully supervised action segmentation make frame-wise predictions~\cite{lea2017temporal,MS-TCN,li2020ms} while trying to model the temporal relationship between the action labels. These approaches usually suffer from over-segmentation. Recent works~\cite{wang2020boundary,ishikawa2020alleviating} try to overcome the over-segmentation problem by finding the action boundaries and temporally smoothing the predictions inside each action segment. But these post-processing approaches still can not guarantee temporal smoothness.

Action segmentation inference is the problem of making segment-wise smooth predictions from frame-wise probabilities given a known grammar of the actions and their average lengths~\cite{richard2018nnviterbi}. The typical inference in action segmentation involves solving an expensive Viterbi-like dynamic programming problem that finds the best action sequence and its corresponding lengths.
In the literature usually weakly supervised action segmentation approaches~\cite{hildecviu,richard2017weakly,richard2018nnviterbi,CDFL,mucon} use inference at test time. Despite being very useful for action segmentation, the inference problem remains the main computational bottleneck in the action segmentation pipeline~\cite{mucon}.

In this paper, we propose FIFA, a fast anytime approximate inference procedure that achieves comparable performance with respect to the dynamic programming based Viterbi decoding inference at a fraction of the computational time.
Instead of relying on dynamic programming, we formulate the energy function as an approximate differentiable function of segment lengths parameters and use gradient-descent-based methods to search for a configuration that minimizes the approximate energy function.
Given a transcript of actions and the corresponding initial lengths configuration, we define the energy function as a sum over segment level energies. The segment level energy consists of two terms: a length energy term that penalizes the deviations from a global length model and an observation energy term that measures the compatibility between the current configuration and the predicted frame-wise probabilities. A naive approach to model the observation energy would be, to sum up the negative log probabilities of the action labels that are defined based on the length configuration. Nevertheless, such an approach is not differentiable with respect to the segment lengths.
In order to optimize the energy using gradient descent-based methods, the observation energy has to be differentiable with respect to the segment lengths. To this end, we construct a plateau-shaped mask for each segment which temporally locates the segment within the video. 
This mask is parameterized by the segment lengths, the position in the video, and a sharpness parameter. The observation energy is then defined as a product of a segment mask and the predicted frame-wise negative log probabilities, followed by a sum pooling operation.  Finally, a gradient descent-based method is used to find a configuration for the segment lengths that minimizes the total energy.

FIFA is a general inference approach and can be applied at test time on top of different action segmentation approaches for fast inference.
We evaluate our approach on top of the state-of-the-art methods for weakly supervised temporal action segmentation, weakly supervised action alignment, and fully supervised action segmentation. Results on the Breakfast~\cite{breakfast} and Hollywood extended~\cite{hollywoodextended} datasets show that FIFA achieves state-of-the-art results on most metrics.
Compared to the exact inference using the Viterbi decoding, FIFA is at least 5 times faster.
Furthermore, FIFA is an anytime algorithm which can be stopped after each step of the gradient-based optimization, therefore it provides a better speed vs. accuracy trade-off compared to exact inference.

\section{Related Work}
In this section we highlight relevant works addressing fully and weakly supervised action segmentation that have been recently achieved.

\paragraph{Fully Supervised Action Segmentation.}
In fully supervised action segmentation, frame-level labels are used for training.
Initial attempts for action segmentation applied action classifiers on a sliding window over the video frames~\cite{rohrbach2012database,karaman2014fast}. However, these approaches did not capture the dependencies between the action segments. With the objective of capturing the context over long video sequences, context free grammars~\cite{vo2014stochastic,pirsiavash2014parsing} 
or hidden Markov models (HMMs)~\cite{lea2016segmental,kuehne2016end,kuehne2020hybrid} are typically combined with frame-wise classifiers. Recently, temporal convolutional networks showed good performance for the temporal action segmentation task using encoder-decoder architectures~\cite{lea2017temporal,lei2018temporal} or even multi-stage architectures~\cite{MS-TCN,li2020ms}. Many approaches further improve the multi-stage architectures by applying post-processing based on boundary-aware pooling operation~\cite{wang2020boundary,ishikawa2020alleviating} or graph-based reasoning~\cite{huang2020improving}.
Without any inference most of the fully-supervised approaches suffer from oversegmentation at test time.

\paragraph{Weakly Supervised Action Segmentation.}
To reduce the annotation cost, many approaches that rely on a weaker form of supervision have been proposed. Earlier approaches
apply discriminative clustering to align video frames to movie scripts~\cite{duchenne2009automatic}. Bojanowski~\etal~\cite{bojanowski2014weakly} proposed to use as supervision the transcripts in the form of ordered lists of actions. Indeed, many approaches rely on this form of supervision to train a segmentation model using connectionist temporal 
classification~\cite{huang2016connectionist}, dynamic time warping~\cite{d3tw} or energy-based learning~\cite{CDFL}. 
In~\cite{isba}, an iterative training procedure is used to refine the transcript. A soft labeling mechanism is further applied at the boundaries between action segments. 
Kuehne~\etal~\cite{kuehne2017weakly} applied a speech recognition system based on a HMM and Gaussian mixture model (GMM) to align video frames to transcripts. The approach generates pseudo ground truth labels for the training videos and iteratively refine them. A similar idea has been recently used in~\cite{richard2017weakly,kuehne2020hybrid}. 
Richard~\etal~\cite{richard2018nnviterbi} combined the frame-wise loss function with the Viterbi algorithm to generate the target labels. At inference time, these approaches iterate over the training transcripts and select the one that matches best the testing video. By contrast, Souri~\etal~\cite{mucon} predict the transcript besides the frame-wise scores at inference time.
State of the art weakly supervised action segmentation approaches require time consuming dynamic programming based inference at test time.

\paragraph{Energy-Based Inference.} In energy-based inference methods, gradient descent is used at inference time as described in~\cite{lecun2006tutorial}. The goal is to minimize an energy function that measures the compatibility between the input variables and the predicted variables. This idea has been exploited for many structured prediction tasks such as image generation~\cite{gatys2015neural,johnson2016perceptual}, machine translation~\cite{hoang2017towards} and structured prediction energy networks~\cite{belanger2017end}. Belanger and McCallum~\cite{belanger2016structured} relaxed the discrete output space for multi-label classification tasks to a continuous space and used gradient descent to approximate the solution. Gradient-based methods have also been used for other applications such as generating adversarial examples~\cite{goodfellow2014explaining} and learning text embeddings~\cite{le2014distributed}.

\section{Background} \label{sec:background}
The following sections introduce all the concepts and notations required to understand the proposed FIFA methodology.

\subsection{Action Segmentation} \label{sec:background:action_segmentation}
In action segmentation, we want to temporally localize all the action segments occurring in a video. In this paper, we consider the case where the actions are from a predefined set of $M$ classes (a background class is used to cover uninteresting parts of a video).
The input video of length $T$ is usually represented as a set of $d$ dimensional features vectors $x_{1:T} = (x_1, \dots, x_T)$. These features are extracted offline and are assumed to be the input to the action segmentation model.
The output of action segmentation can be represented in two ways:
\begin{itemize}
    \item Frame-wise representation $y_{1:T} = (y_1, \dots, y_T)$ where  $y_t$ represents the action label at time $t$.
    \item Segment-wise representation $s_{1:N} = (s_1, \dots, s_N)$ where segment $s_n$ is represented by both the action label of the segment $c_n$ and its corresponding length $\ell_n$ i.e. $s_n = (c_n, \ell_n)$. The ordered list of actions $c_{1:N}$ is usually referred to as the \textit{transcript}.
\end{itemize}
These two representations are equal and redundant i.e. it is possible to compute one from the other.
In order to transfer from the segment-wise to the frame-wise representation, we introduce a mapping $\alpha(t; c_{1:N}, \ell_{1:N})$ which outputs the action label at frame t given the segment-wise labeling.

The target labels to train a segmentation model, depend on the level of supervision.
In fully supervised action segmentation \cite{MS-TCN,li2020ms,wang2020boundary}, the target label for each frame is provided.
However, in weakly supervised approaches \cite{richard2018nnviterbi,CDFL,mucon} only the ordered list of action labels are provided during training while their lengths are unknown.

Recent fully supervised approaches for action segmentation like MSTCN \cite{MS-TCN} and its variants directly predict the frame-wise representation $y_{1:T}$ by choosing the action label with the highest probability for each frame independently. This results in predictions that are sometimes oversegmented.

Conversely, recent weakly supervised action segmentation approaches like NNV \cite{richard2018nnviterbi} and follow-up work include an inference stage during testing where they explicitly predict the segment-wise representation.
This inference stage involves a dynamic programming algorithm for solving an optimization problem which is a computational bottleneck for these approaches.

\subsection{Inference in Action Segmentation} \label{sec:background:inference}
During testing, the inference stage involves an optimization problem to find the most likely segmentation for the input video i.e.,
\begin{align}
    c_{1:N}, \ell_{1:N} = \underset{\hat{c}_{1:N}, \hat{\ell}_{1:N}}{\mathrm{argmax}} \Big\{ p(\hat{c}_{1:N}, \hat{\ell}_{1:N} | x_{1:T}) \Big\}.
\end{align}

Given the transcript $c_{1:N}$, the inference stage boils down to finding the segment lengths $\ell_{1:N}$ by aligning the transcript to the input video i.e.,
\begin{align}
    \ell_{1:N} = \underset{\hat{\ell}_{1:N}}{\mathrm{argmax}} \Big\{ p(\hat{\ell}_{1:N} | x_{1:T}, c_{1:N}) \Big\}.
    \label{eq:inference}
\end{align}
In approaches like NNV \cite{richard2018nnviterbi} and CDFL \cite{CDFL}, the transcript is found by iterating over the transcripts seen during training and selecting the transcript that achieves the most likely alignment by optimizing (\ref{eq:inference}). In MuCon \cite{mucon}, the transcript is predicted by a sequence to sequence network.

The probability defined in (\ref{eq:inference}) is broken down by making independences assumption between frames 
\begin{equation}
    \begin{aligned}
        p(\hat{\ell}_{1:N} | x_{1:T}, c_{1:N}) = \prod_{t=1}^{T} p\big( \alpha(t; c_{1:N}, \hat{\ell}_{1:N}) | x_t \big)
                                                    \cdot \prod_{n=1}^{N} p\big( \hat{\ell}_n | c_n \big) 
        \label{eq:prob_decomp}
    \end{aligned}
\end{equation}
where $p\big( \alpha(t)| x_t \big)$ is referred to as the observation model and $p\big( \ell_n | c_n \big)$ as the length model. Here $\alpha(t)$ is the mapping from time $t$ to the action label given the segmentwise labeling.
The observation model estimates the frame-wise action probabilities and is implemented using a neural network.
The length model is used to constrain the inference defined in (\ref{eq:inference}) with the assumption that the length of segments for the same action follow a particular probability distribution.  
The segment length is usually modelled by a Poisson distribution with a class dependent mean parameter $\lambda_{c_n}$ i.e.,
\begin{align}
    p\big( \ell_n | c_n \big) = \frac{\lambda_{c_n}^{\ell_n} \mathrm{exp}(- \lambda_{c_n}) }{\ell_n!}.
    \label{eq:poisson}
\end{align}

This optimization is solved using an expensive dynamic programming based Viterbi decoding ~\cite{richard2018nnviterbi}. For details on how to solve this optimization problem using Viterbi decoding please refer to the supplementary material.


\section{FIFA: Fast Inference Approximation}

\begin{figure*}[t]
   \centering
      \includegraphics[width=0.9\linewidth]{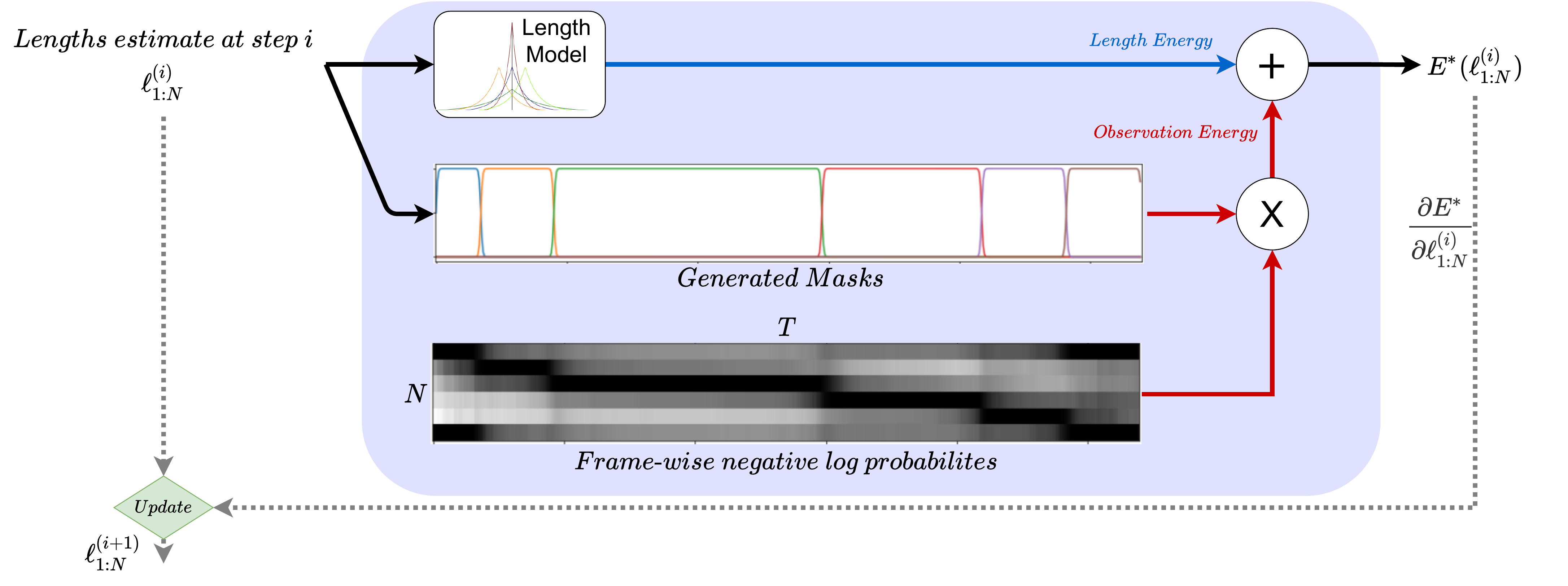}
      \caption{Overview of the FIFA optimization process. At each step in the optimization, using the current length estimates a set of masks are generated. Using the generated masks and the frame-wise negative log probabilities, the observation energy is calculated in an approximate but differentiable manner. The length energy is calculated from the current length estimate and added to the observation energy to calculate the total energy value. Taking the gradient of the total energy with respect to the length estimates we can update it using a gradient step.}
   \label{fig:fifa}
\end{figure*}

Our goal is to introduce a fast inference algorithm for action segmentation.
We want the fast inference to be applicable in both weakly supervised and fully supervised action segmentation.
We also want the fast inference to be flexible enough to work with different action segmentation methods.
To this end, we introduce FIFA, a novel approach for fast inference for action segmentation.

In the following for brevity we write the mapping $\alpha(t; c_{1:N}, \ell_{1:N})$ simply as $\alpha(t)$.
Maximizing probability (\ref{eq:inference}) can be rewritten as minimizing the negative log of that probability
\begin{equation}
    \begin{aligned}
         \mathrm{argmax} \bigg\{ p(\hat{\ell}_{1:N} | x_{1:T}, c_{1:N}) \bigg\} =
         \mathrm{argmin} \bigg\{- \log \big( p(\hat{\ell}_{1:N} | x_{1:T}, c_{1:N}) \big) \bigg\}
    \end{aligned}
\end{equation}
which we refer to as the energy $E(\ell_{1:N})$.
Using (\ref{eq:prob_decomp}) the energy is rewritten as
\begin{equation}
    \label{eq:energy:first}
    \begin{aligned}
        E(\ell_{1:N}) = - \log \bigg(p(\ell_{1:N} | x_{1:T}, c_{1:N}) \bigg)
                      =& - \log \bigg(\prod_{t=1}^{T} p\big( \alpha(t) | x_t \big)
                                    \cdot \prod_{n=1}^{N} p\big( \ell_n | c_n \big) \bigg) \\
                      =& \underbrace{\sum_{t=1}^{T} - \log p\big( \alpha(t) | x_t \big)}_{\textstyle{E_{o}}}
                                    + \underbrace{\sum_{n=1}^{N} - \log p\big( \ell_n | c_n \big)}_{\textstyle{E_{\ell}}}.
    \end{aligned}
\end{equation}

The first term in (\ref{eq:energy:first}), $E_o$ is referred to as the observation energy. This term calculates the cost of assigning the labels for each frame and will be calculated from the frame-wise probability estimates.
The second term $E_\ell$ is referred to as the length energy. This term is the cost of each segment having a length given that we assume some average length for actions of a specific class.

We proposed to optimize the energy defined in (\ref{eq:energy:first}) using gradient based optimization in order to avoid the need for time-consuming dynamic programming. We start with an initial estimate of the lengths (obtained from the length model of each approach or calculated from training data when available) and update our estimate to minimize the energy function.

As the energy function $E(\ell_{1:N})$ is not differentiable with respect to the lengths, we have to calculate a relaxed and approximate energy function $E^*(\ell_{1:N})$ that respects this mathematical property.

\subsection{Approximate Differentiable Energy $E^*$}
The energy function $E$ as defined in (\ref{eq:energy:first}) is not differentiable in two parts.
First the observation energy term $E_o$ is not differentiable because of the $\alpha(t)$ function.
Second, the length energy term $E_\ell$ is not differentiable because it expects natural numbers as input and cannot be computed on real values which are dealt with in gradient-based optimization.
Below we describe how we approximate and make each of the terms differentiable.

\subsubsection{Approximate Differentiable Observation Energy}

Consider a $N \times T$ matrix $P$ containing negative log probabilities, i.e.
\begin{align}
    P[n, t] = -\log p(c_n|x_t).
\end{align}
Imagine a mask matrix $M$ with the same size $N \times T$ where
\begin{align}
    M[n, t] = \begin{cases}
        0 & if ~ \alpha(t) \ne c_n \\
        1 & if ~ \alpha(t) = c_n
    \end{cases}.
\end{align}
Using the mask matrix we can rewrite the observation energy term as
\begin{align}
    E_o = \sum_{t=1}^{T} \sum_{n=1}^{N} M[n, t] \cdot P[n, t].
    \label{eq:eng_obs:exact}
\end{align}

In order to make the observation energy term differentiable with respect to the length, we propose to construct an approximate differentiable mask matrix $M^*$.
We use the following smooth and parametric plateau function
\begin{align}
    f(t|\lambda^c, \lambda^w, \lambda^s) = \frac{1}{(e^{\lambda^s(t-\lambda^c-\lambda^w)} + 1)(e^{\lambda^s(-t+\lambda^c-\lambda^w)} + 1)}
\end{align}
from \cite{moltisanti2019action}. This plateau function has three parameters and it is  differentiable with respect to them: $\lambda^c$ controls the center of the plateau, $\lambda^w$ is the width and $\lambda^s$ is the sharpness of the plateau function.

While the sharpness of the plateau functions $\lambda^s$ used to construct the approximate mask $M^*$ is fixed as a hyper-parameter of our approach, the center $\lambda^c$ and the width $\lambda^w$ are computed from the lengths $\ell_{1:N}$.
First we calculate the starting position of each plateau function $b_n$ as
\begin{align}
    b_1 = 0, b_n = \sum_{n' = 1}^{n-1} \ell_{n'}.
\end{align}
We can then define both the center and the width parameters of each plateau function as
\begin{equation}
    \begin{aligned}
        \lambda_n^c &= b_n + \ell_n / 2, ~~~
        \lambda_n^w &= \ell_n / 2
    \end{aligned}
\end{equation}
and define each row of the approximate mask as
\begin{align}
    M^*[n, t] = f(t| \lambda_n^c, \lambda_n^w, \lambda^s).
\end{align}
Now we can calculate a differentiable approximate observation energy similar to (\ref{eq:eng_obs:exact}) as
\begin{align}
    E^*_o = \sum_{t=1}^{T} \sum_{n=1}^{N} M^*[n, t] \cdot P[n, t].
\end{align}

\subsubsection{Approximate Differentiable Length Energy}
For the gradient-based optimization, we must relax the length values to be positive real values instead of natural numbers.
As the Poisson distribution (\ref{eq:poisson}) is only defined on natural numbers, we propose to use a substitute distribution defined on real numbers.
As a replacement, we experiment with a Laplace distribution and a Gaussian distribution. In both cases, the scale or the width parameter of the distribution is assumed to be fixed.

We can rewrite the length energy $E_\ell$ as the approximate length energy
\begin{equation}
    \begin{aligned}
        E^*_\ell(\ell_{1:N}) = \sum_{n=1}^{N} - \log p(\ell_n|\lambda^{\ell}_{c_n}),
    \end{aligned}
\end{equation}
where $\lambda^{\ell}_{c_n}$ is the expected value for the length of a segment from the action $c_n$.
In case of the Laplace distribution this length energy will be
\begin{equation}
    \begin{aligned}
        E^*_\ell(\ell_{1:N}) = \frac{1}{Z} \sum_{n=1}^{N} | \ell_n -  \lambda^{\ell}_{c_n}|,
    \end{aligned}
\end{equation}
where $Z$ is the constant normalization factor.
This means that the length energy will penalize any deviation from the expected average length linearly. Similarly, for the Gaussian distribution, the length energy will be
\begin{equation}
    \begin{aligned}
        E^*_\ell(\ell_{1:N}) = \frac{1}{Z} \sum_{n=1}^{N} | \ell_n -  \lambda^{\ell}_{c_n}|^2,
    \end{aligned}
\end{equation}
which means that the Gaussian length energy will penalize any deviation from the expected average length quadratically.

With the objective to maintain a positive value for the length during the optimization process, we estimate the length in log space and convert it to absolute space only in order to compute both the approximate mask matrix $M^*$ and the approximate length energy $E^*_\ell$.

\subsubsection{Approximate Energy Optimization}
The total approximate energy function is defined as a weighted sum of both the approximate observation and the approximate length energy functions
\begin{align}
    E^*(\ell_{1:N}) = E^*_o(\ell_{1:N}, Y) + \beta E^*_\ell(\ell_{1:N})
\end{align}
where $\beta$ is the multiplier for the length energy.

Given an initial length estimate $\ell^0_{1:N}$, we iteratively update this estimate to minimize the total energy. Figure~\ref{fig:fifa} illustrates the optimization step for our approach. During each optimization step, we first calculate the energy $E^*$ and then calculate the gradients of the energy with respect to the length values. Using the calculated gradients, we update the length estimate using a gradient descent update rule such as SGD or Adam. After a certain number of gradient steps (50 steps in our experiments) we will finally predict the segment length.

During testing, if the transcript is provided then it is used (e.g. using the MuCon \cite{mucon} approach or in a weakly supervised action alignment setting). However, if the latter is not known (e.g. in a fully supervised approach or CDFL \cite{CDFL} for weakly supervised action segmentation) we perform the optimization for each of the transcripts seen during training and select the most likely one based on the final energy value at the end of the optimization.

The initial length estimates are calculated from the length model of each approach in case of weakly supervised setting whereas in fully supervised setting the average length of each action class is calculated from the training data and used as the initial length estimates.
The initial length estimates are also used as the expected length parameters for the length energy calculations.

The hyper-parameters like the choice of the optimizer, number of steps, learning rate, and the mask sharpness, remain as the hyper-parameters of our approach.

\section{Experiments}


\begin{figure}[tb]
   \begin{minipage}{0.48\textwidth}
      \centering
         \includegraphics[width=0.95\columnwidth]{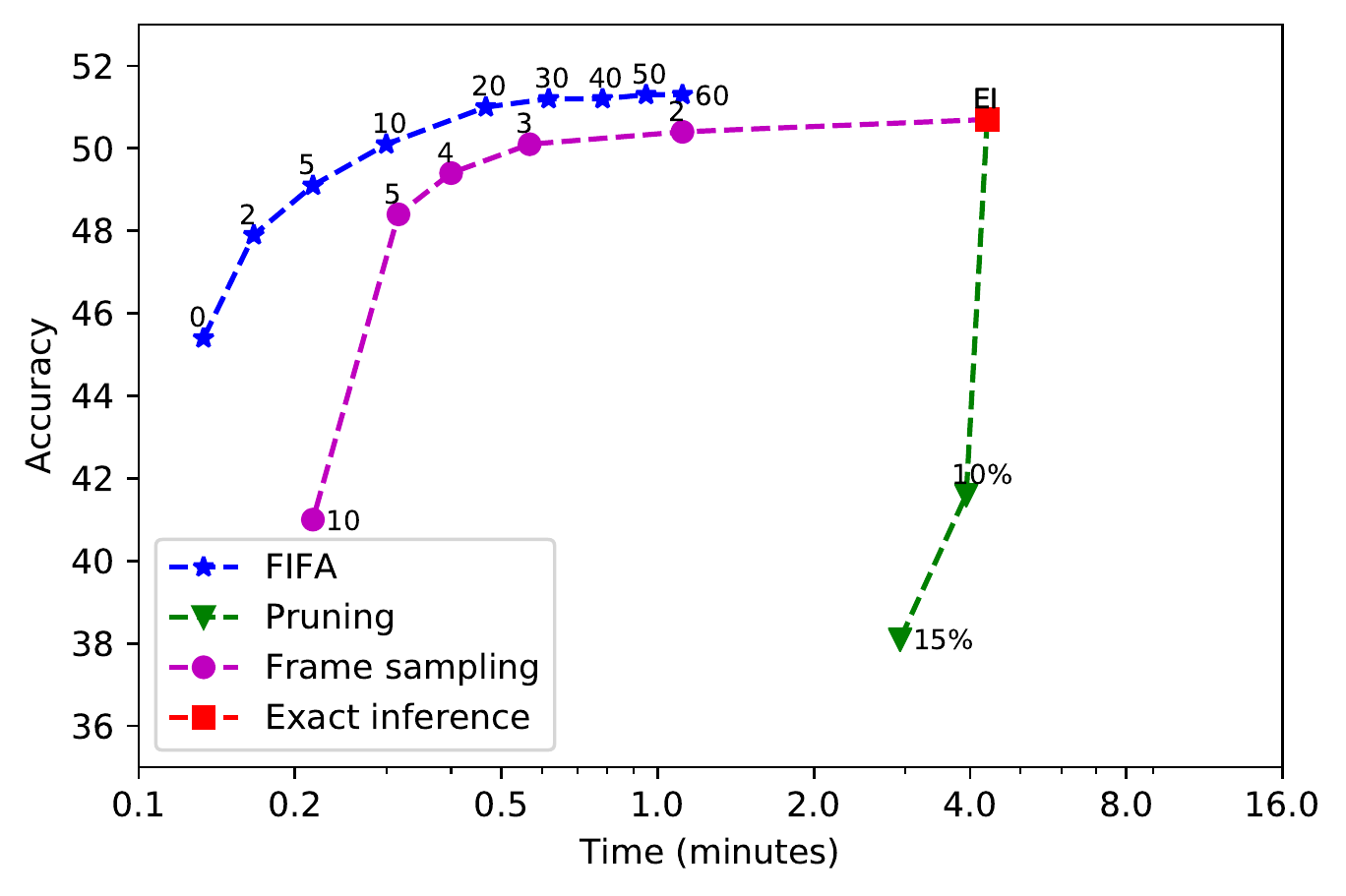}
      \caption{
         Speed vs. accuracy trade-off of different inference approaches applied to the MuCon method.
         Using FIFA we can achieve a better speed vs. accuracy trade-off compared to frame sampling or hypothesis pruning in exact inference.}
      \label{fig:speed_acc}
   \end{minipage}\hfill
   \begin{minipage}{0.48\textwidth}
      \centering
         \includegraphics[width=0.95\columnwidth]{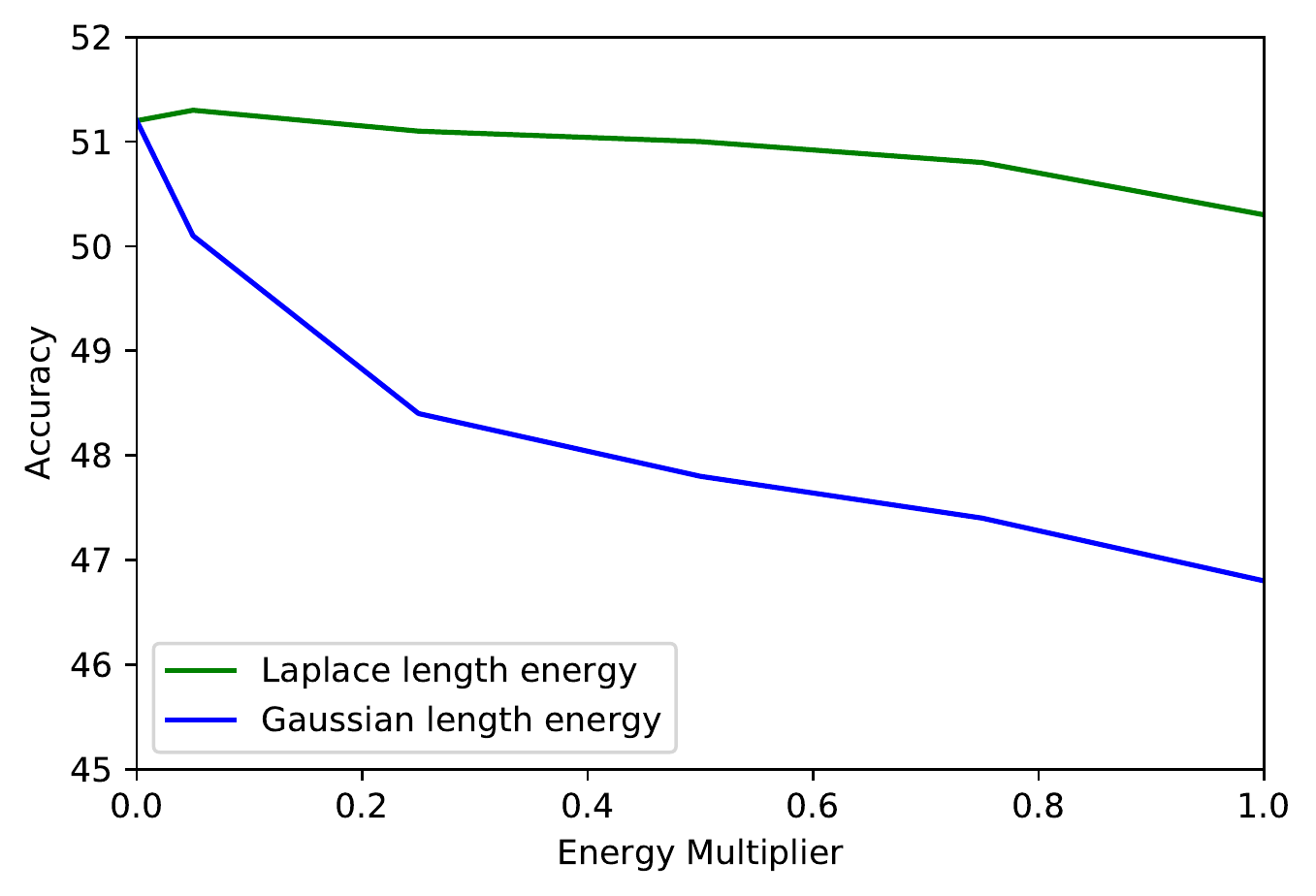}
      \caption{Effect of the length energy multiplier for Laplace and Gaussian length energy. Accuracy is calculated on the breakfast dataset using FIFA applied to the MuCon approach trained in the weakly supervised action segmentation setting.}
      \label{fig:len_energy}
   \end{minipage}
\end{figure}

\subsection{Evaluation Protocols and Datasets}
We evaluate FIFA on 3 different tasks: weakly supervised action segmentation, fully supervised action segmentation, and weakly supervised action alignment. Results for action alignment are included in the supplementary material.
We obtain the source code for the state-of-the-art approaches on each of these tasks and train a model using the standard training configuration of each model.
Then we apply FIFA as a replacement for an existing inference stage or as an additional inference stage.

We evaluate our model using the Breakfast~\cite{breakfast} and Hollywood extended~\cite{hollywoodextended} datasets on the 3 different tasks. Details of the datasets are included in the supplementary material.

\subsection{Results and Discussions}
In this section, we study the speed-accuracy trade-off and the impact of the length model. 
Additional ablation experiments are included in the supplementary material.
\paragraph{Speed vs. Accuracy Trade-off.}
One of the major benefits of FIFA is the flexibility of choosing the number of optimization steps. The number of steps of the optimization can be a tool to trade-off speed vs. accuracy.
In exact inference, we can use frame-sampling i.e. lowering the resolution of the input features, or hypothesis pruning i.e. beam search for speed vs. accuracy trade-off.

Figure~\ref{fig:speed_acc} plots the speed vs. accuracy trade-off of exact inference compared to FIFA.
We observe that FIFA provides a much better speed-accuracy trade-off as compared to frame-sampling for exact inference. The best performance after 50 steps with $5.9\%$ improvement on the MoF accuracy compared to not performing any inference.

\paragraph{Impact of the Length Energy.}
For the length energy, we assume that the segment lengths follow a Laplace distribution. Figure~\ref{fig:len_energy} shows the impact of the length energy multiplier on the performance. While the best accuracy is achieved with a multiplier of $0.05$, our approach is robust to the choice of these hyper-parameters. We further experimented with a Gaussian length energy. However, as shown in the figure, the performance is much worse compared to the Laplace energy. This is due to the quadratic penalty that dominates the total energy, which makes the optimization biased towards the initial estimate and ignores the observation energy.  


\paragraph{Impact of Length Model Initialization.}
Since FIFA starts with an initial estimate for the lengths, the choice of initialization might have an impact on the performance. Table~\ref{tab:initialization} shows the effect of initializing the lengths with equal values compared to using the length model of MuCon~\cite{mucon} for the weakly supervised action segmentation on the Breakfast dataset. As shown in the table, FIFA is more robust to initialization compared to the exact inference as the drop in performance is approximately half of the exact inference.

\begin{table}[tb]
   \centering
   \resizebox{.45\columnwidth}{!}{%
      \begin{tabular}{clc}
         \toprule
         Inference Method & initialization & MoF   \\
         \midrule
           Exact & MuCon~\cite{mucon}    & \textbf{50.7}  \\
                 & Equal  &  48.8 (-1.9) \\
         \midrule
           FIFA   & MuCon~\cite{mucon}   &  \textbf{51.3} \\
                  & Equal &  50.2 (-1.1) \\
         \bottomrule
      \end{tabular}
    }
   \caption{Impact of the Length Model initialization for MuCon using exact inference and FIFA for weakly supervised action segmentation on the Breakfast dataset.}
   \label{tab:initialization}
\end{table}

\begin{figure*}[t]
    \centering
       \includegraphics[width=0.9\linewidth]{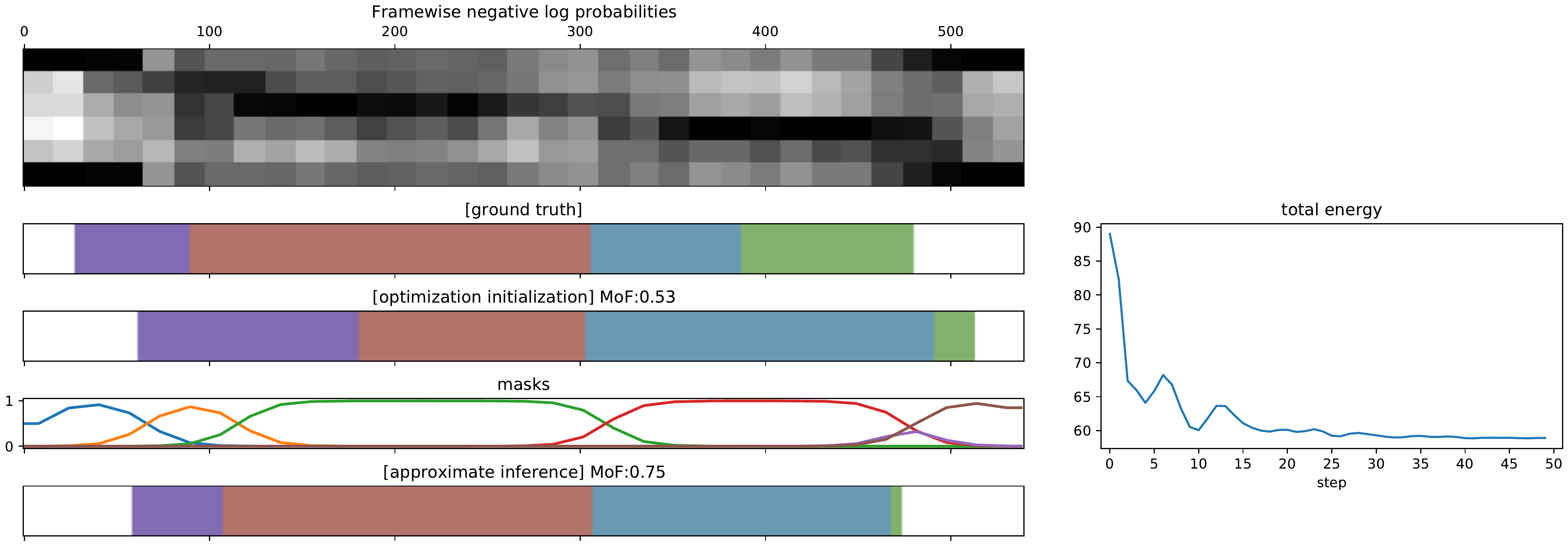}
    \caption{Visualization of the FIFA optimization process. On the right the values of the total approximate energy is plotted.
    On the left, negative log probability values, ground truth segmentation, optimization initialization, the masks and the segmentation after inference is ploted.}
    \label{fig:q_res}
 \end{figure*}

\subsection{Comparison to State of the Art}

%

In this section, we compare FIFA to other state-of-the-art approaches.

\paragraph{Weakly Supervised Action Segmentation.}
We apply FIFA on top of two state-of-the-art approaches for weakly supervised action segmentation namely MuCon~\cite{mucon} and CDFL~\cite{CDFL} on the Breakfast dataset~\cite{breakfast} and report the results in Table\ref{tab:weak_segmentation}. FIFA applied on CDFL achieves a 12 times faster inference speed while obtaining results comparable to exact inference.
FIFA applied to MuCon achieves a 5 times faster inference speed and obtains a new state-of-the-art performance on the Breakfast dataset on most of the metrics.

\begin{table}
   \begin{minipage}{0.48\textwidth}
      \centering
      \resizebox{\columnwidth}{!}{%
         \begin{tabular}{lccccc}
            \toprule
            Method & MoF & MoF-BG & IoU & IoD & Time (min)  \\
            \midrule
            ISBA~\cite{isba} & 38.4 & 38.4 & 24.2 & 40.6 & 0.01 \\
            NNV~\cite{richard2018nnviterbi} & 43.0 & - & - & - & 234 \\
            D3TW~\cite{d3tw} & 45.7 & - & - & - & - \\
            \midrule
            CDFL~\cite{CDFL} & 50.2 & 48.0 & 33.7 & 45.4 & - \\
            CDFL$^*$ & 49.4 & 47.5 & 35.2 & 46.4 & 260 \\
            FIFA + CDFL$^*$  & 47.9  & 
                              46.3  & 
                              34.7  & 
                              48.0  &   
                              20.4 ($\times$12.8)\\
            \midrule
            MuCon~\cite{mucon} & 47.1 & - & - & - & - \\
            MuCon$^*$ & \underline{50.7} & \underline{50.3} & \underline{40.9} & \textbf{54.0} & 4.1 \\
            FIFA + MuCon$^*$  & \textbf{51.3}  & 
                              \textbf{50.7}  & 
                              \textbf{41.1}  & 
                              \underline{53.3}   & 
                              0.8 ($\times$5.1)\\
            \bottomrule
         \end{tabular}
      }
      \caption{Results for weakly supervised action segmentation on the Breakfast dataset. $^*$ indicates results obtained by running the code on our machine.} 
      \label{tab:weak_segmentation}
   \end{minipage}\hfill
   \begin{minipage}{0.48\textwidth}
      \centering
      \resizebox{\columnwidth}{!}{%
         \begin{tabular}{lccccc}
            \toprule
            Method & \multicolumn{3}{c}{F1@\{10, 25, 50\}} & Edit & MoF  \\
            \midrule
            BCN~\cite{wang2020boundary} & 68.7 & 65.5 & \underline{55.0} & 66.2 & \textbf{70.4}  \\
            ASRF~\cite{ishikawa2020alleviating} & \underline{74.3} & 68.9 & \textbf{56.1} & 72.4 & 67.6  \\
            \midrule
            MS-TCN++~\cite{li2020ms} & 64.1 & 58.6 & 45.9 & 65.6 & 67.6  \\
            FIFA + MS-TCN++ & \underline{74.3} & \underline{69.0} & 54.3 & \underline{77.3} & 67.9 \\
            \midrule
            MS-TCN~\cite{MS-TCN} & 52.6 & 48.1 & 37.9 & 61.7 & 66.3  \\
            FIFA + MS-TCN   & \textbf{75.5} & \textbf{70.2} & 54.8 & \textbf{78.5} & \underline{68.6}\\
            \bottomrule
         \end{tabular}
      }
      \caption{Results for fully supervised action segmentation setup on the Breakfast dataset.}
      \label{tab:full_segmentation}
   \end{minipage}
\end{table}


Similarly for the Hollywood extended dataset~\cite{hollywoodextended} we apply FIFA to MuCon~\cite{mucon} and report the results in Table~\ref{tab:weak_hollywood}.
FIFA applied on MuCon achieves a 4 times faster inference speed while obtaining results comparable to exact inference.

\paragraph{Fully Supervised Action Segmentation.}
In the fully supervised action segmentation we apply FIFA on top of MS-TCN~\cite{MS-TCN} and its variant MS-TCN++~\cite{li2020ms} on the Breakfast dataset~\cite{breakfast} and report the results in Table~\ref{tab:full_segmentation}.
MS-TCN and MS-TCN++ are approaches that do not perform any inference at test time. This usually results in over-segmentation and low F1 and Edit scores.
Applying FIFA on top of these approaches improves the F1 and Edit scores significantly.
FIFA applied on top of MS-TCN achieves state-of-the-art performance and sets new state-of-the-art performance on most metrics.

For the Hollywood extended dataset~\cite{hollywoodextended}, we train MS-TCN~\cite{MS-TCN} and report results comparing exact inference (EI) compared to FIFA in Table~\ref{tab:full_hollywood}. We observe that MS-TCN using an inference algorithm achieves new state-of-the-art results on this dataset. FIFA is comparable or better than exact inference on this dataset.


\begin{table}
   \begin{minipage}{0.48\textwidth}
      \centering
      \resizebox{\columnwidth}{!}{%
         \begin{tabular}{lcccl}
            \toprule
            Method & MoF-BG & IoU & Time (speedup)\\
            \midrule
            ISBA~[6] & 34.5 & 12.6 & -  \\
            D3TW~[x] & 33.6  &   & -  \\
            CDFL~[x] & 40.6  & 19.5 &  - \\
            MuCon~[33]  & 41.6  &   & - \\
            \midrule
            MuCon$^*$  & 40.1  & 13.9  & 53 \\
            MuCon + FIFA$^*$  & 41.2  & 13.7  & 13 ($\times$4.1) \\
            \bottomrule
         \end{tabular}
      }
      \caption{Results for weakly supervised action segmentation on the Hollywood extended dataset. Time is reported in seconds. $^*$ indicates results obtained by running the code on our machine.}
      \label{tab:weak_hollywood}
   \end{minipage}\hfill
   \begin{minipage}{0.48\textwidth}
      \centering
      \resizebox{\columnwidth}{!}{%
         \begin{tabular}{lcccc}
            \toprule
            Method & MoF & MoF-BG & IoU & IoD \\
            \midrule
            HTK~[19] & 39.5 &   & 8.4 &  \\
            ED-TCN~[6] & 36.7 & 27.3 & 10.9 & 13.1 \\
            ISBA~[6] & 54.8 & 33.1 & 20.4 & 28.8\\
            \midrule
            MSTCN~[1] (+ EI)  & 64.9 & 35.0 & 22.6 & 33.2 \\
            MSTCN + FIFA  & 66.2 & 34.8 & 23.9 & 35.8 \\
            \bottomrule
         \end{tabular}
      }
      \caption{Results for fully supervised action segmentation on the Hollywood extended dataset. EI stands for Exact Inference.}
      \label{tab:full_hollywood}
   \end{minipage}
\end{table}

\subsection{Qualitative Example}


A qualitative example of the FIFA optimization process is depicted in Figure~\ref{fig:q_res}. For further qualitative examples, failure cases, and details please refer to the supplementary material.



\section{Conclusion}

In this paper, we proposed FIFA a fast approximate inference procedure for action segmentation and alignment. Unlike previous methods, the proposed method does not rely on any expensive Viterbi decoding for inference. Instead, FIFA optimizes a differentiable energy function that can be minimized using gradient-descent which allows for a fast but also accurate inference during testing. 
We evaluated FIFA on top of fully and weakly supervised methods trained on the Breakfast dataset. The results show that FIFA is able to achieve comparable or better performance, while being at least 5 time faster than exact inference.
%
%
%

\bibliographystyle{splncs04}
\bibliography{references}

\newpage

\renewcommand\thefigure{\thesection.\arabic{figure}}
\renewcommand\thetable{\thesection.\arabic{table}}
\setcounter{figure}{0} 
\setcounter{table}{0} 

\appendix

\section{Detials of Exact Inference}
The NNV approach \cite{richard2018nnviterbi} proposes an exact solution to the inference problem  using a Viterbi-like dynamic programming method.
This dynamic programming approach was later adopted by CDFL \cite{CDFL} and MuCon \cite{mucon}.
First an auxiliary function $Q(t, \ell, n)$ is defined that yields the best probability score for a segmentation up to frame $t$ satisfying the following conditions:
\begin{itemize}
    \item the length of the last segment is $\ell$,
    \item the last segment was the $n$th segment with label $c_n$.
\end{itemize}
The function $Q$ can be computed recursively. The following two cases are distinguished. The first case defines when no new segment is hypothesized, i.e $\ell > 1$. Then,
\begin{align}
    Q(t, \ell, n) = Q(t-1, \ell-1, n) \cdot p(c_n | x_t),
\end{align}
with the current frame probability being multiplied with the value of the auxiliary function at the previous frame. The second case is a new segment being hypothesized at frame $t$, i.e. $\ell = 1$. Then,
\begin{equation}
\begin{aligned}
    &Q(t, \ell=1, n) = \\
    &\underset{\hat{\ell}} {\mathrm{max}} \bigg\{ Q(t-1, \hat{\ell}, n-1) \cdot p(c_n|x_t) \cdot p(\hat{\ell} | c_{n-1})) \bigg\},
    \label{eq:exact_inf:case2}
\end{aligned}
\end{equation}
where the optimization being calculated over all possible previous segments with length $\hat{\ell}$ and label $c_{n-1}$. Here the probability of the previous segment having length $\hat{\ell}$ and label $c_{n-1}$ is being multiplied to the previous value of the auxiliary function.

The most likely alignment is given by
\begin{align}
    \underset{\ell}{max} \bigg\{ Q(T, \ell, N) \cdot p(\ell|c_N) \bigg\}.
\end{align}
The optimal lengths can be obtained by keeping track of the maximizing arguments $\hat{\ell}$ from (\ref{eq:exact_inf:case2}).

\section{Time Complexity Comparison}

\subsection{Time Complexity of Exact Inference}
The time complexity of the above exact inference is quadratic in the length of the video $T$ and linear in the number of segments $N$. As input videos for action segmentation are usually long, it becomes computationally expensive to calculate.
In practice, \cite{richard2018nnviterbi,CDFL,mucon} limit the maximum size of each segment to a fixed value of $L=2000$. The final time complexity of exact inference is $O(LNT)$.
Furthermore, this optimization process is inherently not parallelizable. This is due to the $\mathrm{max}$ operation in (\ref{eq:exact_inf:case2}).
Experiments have shown \cite{mucon,on_weak_eval} that this inference stage is the main computational bottleneck of action segmentation approaches.

\subsection{Time Complexity of FIFA}
At each optimization step, the time complexity is $O(NT)$, where $N$ is the number of segments and $T$ is the length of the video because we must create the $M^*$ matrix and calculate the element-wise multiplication.
Overall, the FIFA time complexity is $O(MNT)$, where $M$ is the number of optimization steps.
Compared to the exact inference which has a time complexity of $O(LNT)$, where $L$ is the fixed value of 2000, our time complexity is lower since $M$ is usually 50 steps and $N$ is on average 10.

We also want to mention that the proposed approach is inherently a parallelizable optimization method (i.e. values of the mask, the element-wise multiplication, and the calculation of the gradient for each time step can be calculated in parallel) and is independent of any other time step values. This is in contrast to the dynamic programming approaches where the intermediate optimization values for each time step depend on the value of the previous time steps.

\section{Details of the Datasets}
The \textbf{Breakfast} dataset~\cite{breakfast} is the most popular and largest dataset typically used for action segmentation.
It contains more than 1.7k videos of different cooking activities. The dataset consists of 48 different fine-grained actions. In our experiments, we follow the 4 train/test splits provided with the dataset and report the average.

The \textbf{Hollywood extended} dataset~\cite{hollywoodextended} contains 937 videos taken from Hollywood movies. The videos contain 16 different action classes. We follow the train/test split strategy of \cite{isba,richard2018nnviterbi,CDFL}.

The main performance metrics used for weakly supervised action segmentation and alignment are the same as the previous approaches.
The input features are also kept the same depending on the approach we use FIFA with.

\section{Implementation Details}
We implement our approach using the PyTorch\cite{pytorch} library.
For all experiments we set the number of FIFA's gradient-based optimization steps to $50$ and we use the Adam~\cite{adam} optimizer. Mask sharpness and the optimization learning rate is chosen depending on the approach that FIFA is applied on top of.
When applying FIFA on top of MuCon~\cite{mucon} we use $0.3$ as the learning rate and set the mask sharpness to $1.75$. For CDFL~\cite{CDFL}, we set the mask sharpness to $0.1$ and the learning rate to $0.15$. Looking at the visualization in Figure~\ref{fig:q:cdfl:1} it is clear that CDFL provides noisy framewize probability estimates. For this reason a lower mask sharpness is prefered.
When applying FIFA on top of fully supervised approaches like MS-TCN~\cite{MS-TCN} we use mask sharpness value of $15$ and learning rate of $0.02$. Looking at the visualization in Figure~\ref{fig:q:mstcn:1} we see that fully supervised approaches provide clean smooth framewise probabilities and having a sharp mask is recommented in these settings.

\section{Ablation Experiments}

\subsection{Number of Optimization Steps}

In Table~\ref{tab:optimization_steps} we report the results for weakly supervised action segmentation on the Breakfast dataset~\cite{breakfast} using the MuCon\cite{mucon} approach.
The proposed approach achieves the best performance after 50 steps with $5.9\%$ improvement on the MoF accuracy compared to not performing any inference. Moreover, it is more than 5 times faster than the exact inference.

\begin{table}[tb]
      \centering
      \resizebox{0.75\columnwidth}{!}{%
            \begin{tabular}{cccccc }
            \toprule
            Num. Steps & MoF & MoF-BG & IoU & IoD & Time (min) \\
            \midrule
            No inference & 45.4 & 44.7 & 37.3 & 51.2 & 1.0\\
            \midrule
            2 steps  & 47.9 & 47.1 & 39.8 & 53.0 & 1.2 \\
            5 steps  & 49.1 & 48.3 & 40.0 & 52.8 & 1.5 \\
            10 steps & 50.1 & 49.4 & 40.2 & 52.9 & 2.0\\
            30 steps & 51.2 & 50.6 & 41.0 & 53.2 & 4.2\\
            50 steps & \textbf{51.3} & \textbf{50.7} & \textbf{41.1} & \textbf{53.3} & 6.5 \\
            60 steps & \textbf{51.3} & \textbf{50.7} & \textbf{41.1} & \textbf{53.3} & 7.7 \\
            \midrule
            Exact Inference & 50.7 & 50.3 & 40.9 & 54.0 & 32.85 \\
            \bottomrule
            \end{tabular}
            }
      \caption{Impact of the number of optimization steps for FIFA+MuCon for weakly supervised action segmentation on the Breakfast dataset.
      }
      \label{tab:optimization_steps}
\end{table}

\subsection{Optimizer and Its Learning Rate}
\begin{figure}[t]
    \centering
       \includegraphics[width=0.9\columnwidth]{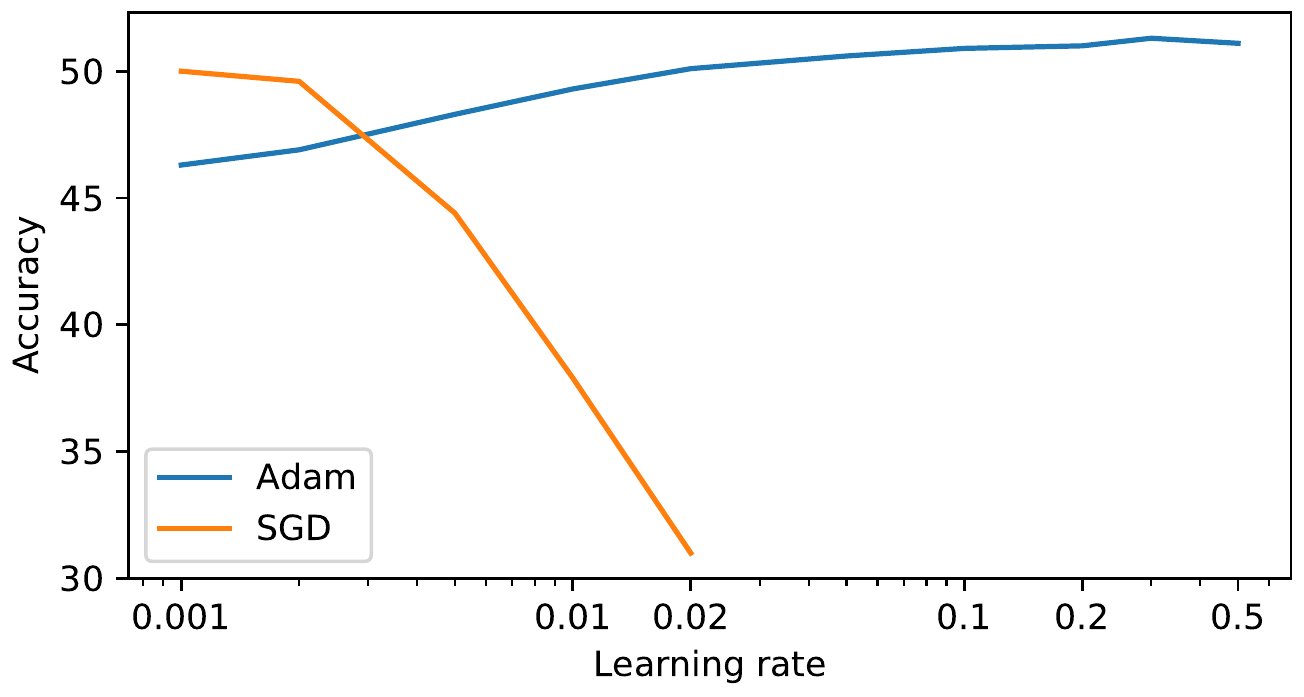}
    \caption{Effect of the learning rate on the performance of weakly supervised action segmentation using FIFA applied on the MuCon approach. Accuracy is calculated on the Breakfast dataset.}
    \label{fig:learning_rate_adam_sgd}
 \end{figure}

The choice of the optimizer used to update the length estimates using the calculated gradients is one of the hyper-parameters of our approach.
We have experimented with two optimizers SGD and Adam.
As shown in Figure~\ref{fig:learning_rate_adam_sgd}, the best performing value for the learning rate hyper-parameter depends on the optimizer used.
For SGD a low value of $0.001$ achieves the best performance with higher values causing major drops in performance.
On the other hand, Adam optimizer works well with a range of learning rate values as it has an internal mechanism to adjust the learning rate. The best performance for Adam is observed at $0.3$.

\begin{figure}[t]
    \centering
       \includegraphics[width=\columnwidth]{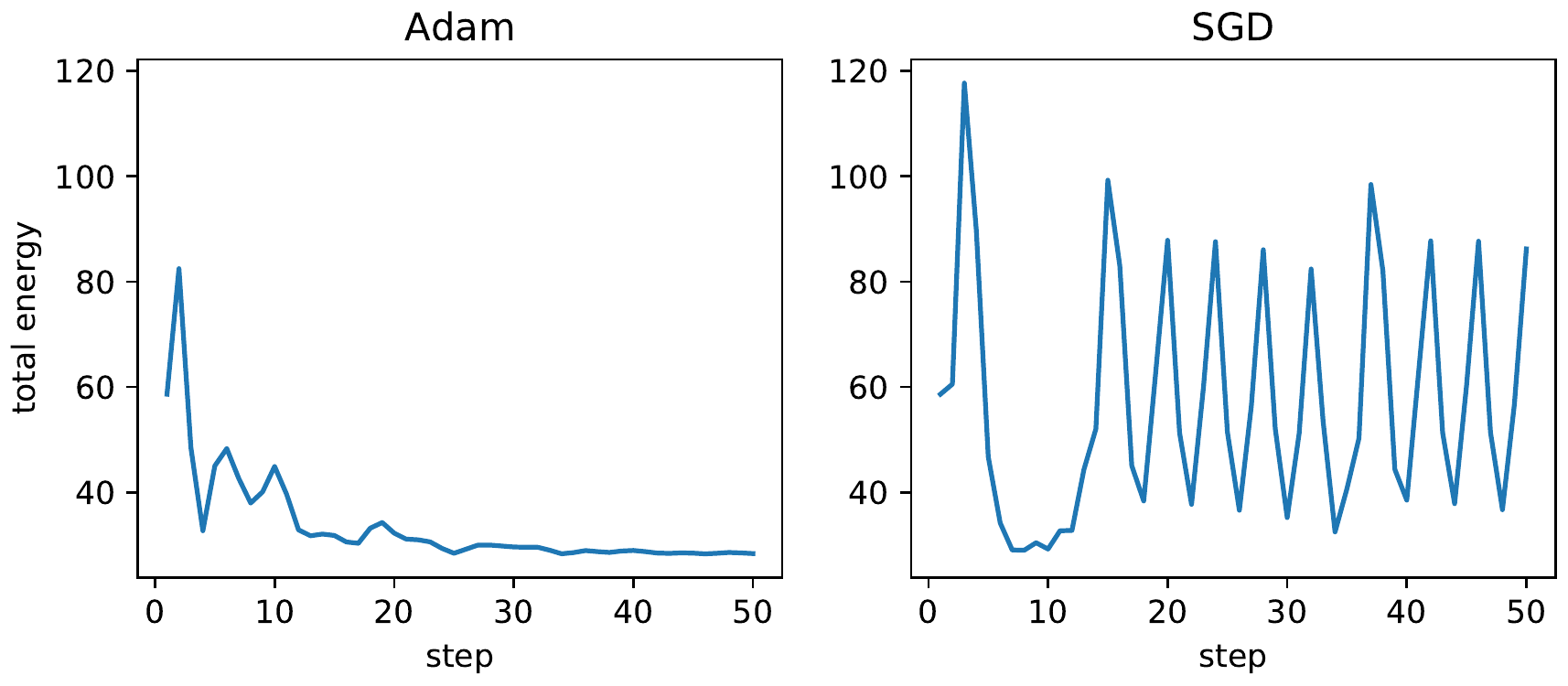}
    \caption{The value of the approximate energy during FIFA optimization for SGD and Adam optimizer for the same inference.}
    \label{fig:energy_adam_vs_sgd}
 \end{figure}

 We further investigate and notice that the reason SGD performs so poorly for large values of the learning rate is that it fluctuates and is not able to optimize the energy effectively. Figure~\ref{fig:energy_adam_vs_sgd} shows the value of the approximate energy during the optimization for Adam and SGD for the same inference.
 We observe that a large learning rate causes SGD to fluctuate while Adam is stable and achieves a lower energy value at the end of the optimization.

\section{Weakly Supervised Action Alignment}
Similar to weakly supervised action segmentation, we apply FIFA on top of CDFL and MuCon for weakly supervised action alignment task on the Breakfast dataset and report the results in Table~\ref{tab:weak_alignment}.
Our experiments show that FIFA applied on top of CDFL achieves state-of-the-art or better than state-of-the-art results on MoF and Mof-BG metrics, whereas FIFA applied on top of MuCon achieves state-of-the-art results for IoD and IoU metrics. 

\begin{table}[tb]
      \centering
      \resizebox{0.75\columnwidth}{!}{%
         \begin{tabular}{lcccc}
            \toprule
            Method & MoF & MoF-BG & IoU & IoD \\
            \midrule
            ISBA~\cite{isba} & 53.5 & 51.7 & 35.3 & 52.3\\
            D$^3$TW~\cite{d3tw} & 57.0 & - & - & 56.3\\
            CDFL~\cite{CDFL} & 63.0 & 61.4 & 45.8 & \underline{63.9}\\
            ADP~\cite{ghoddoosian2021action} & \underline{64.1} & \textbf{65.5} & 43.0 & - \\
            \midrule
            FIFA + CDFL$^*$  & \textbf{65.3} & \underline{64.3} & \underline{46.3} & 61.3\\
            FIFA + MuCon$^*$  & 61.4 & 61.2 & \textbf{48.4} & \textbf{64.1}\\
            \bottomrule
         \end{tabular}
       }
      \caption{Results for weakly supervised action alignment on the Breakfast dataset.}
      \label{tab:weak_alignment}
\end{table}

\section{Qualitative Examples}

In this section we show various qualitative results of applying FIFA for action segmentation.
In each figure on the right, the approximate total energy value is plotted as a function of number of steps.
On the left, at the top, the framewise negative log probabilities ($P$) are visualized.
The ground truth segmentation, optimization initialization, the generated masks and the segmentation obtained after approximate inference using FIFA is visualized in rows 2 to 5.
The MoF metric is also calculated for a single video and reported for the optimization initialization and the approximate decoding.

An animation form of the same figures is also provided in the supplementary material as a single video file.

\begin{figure*}
     \centering
      \includegraphics[width=1.0\linewidth]{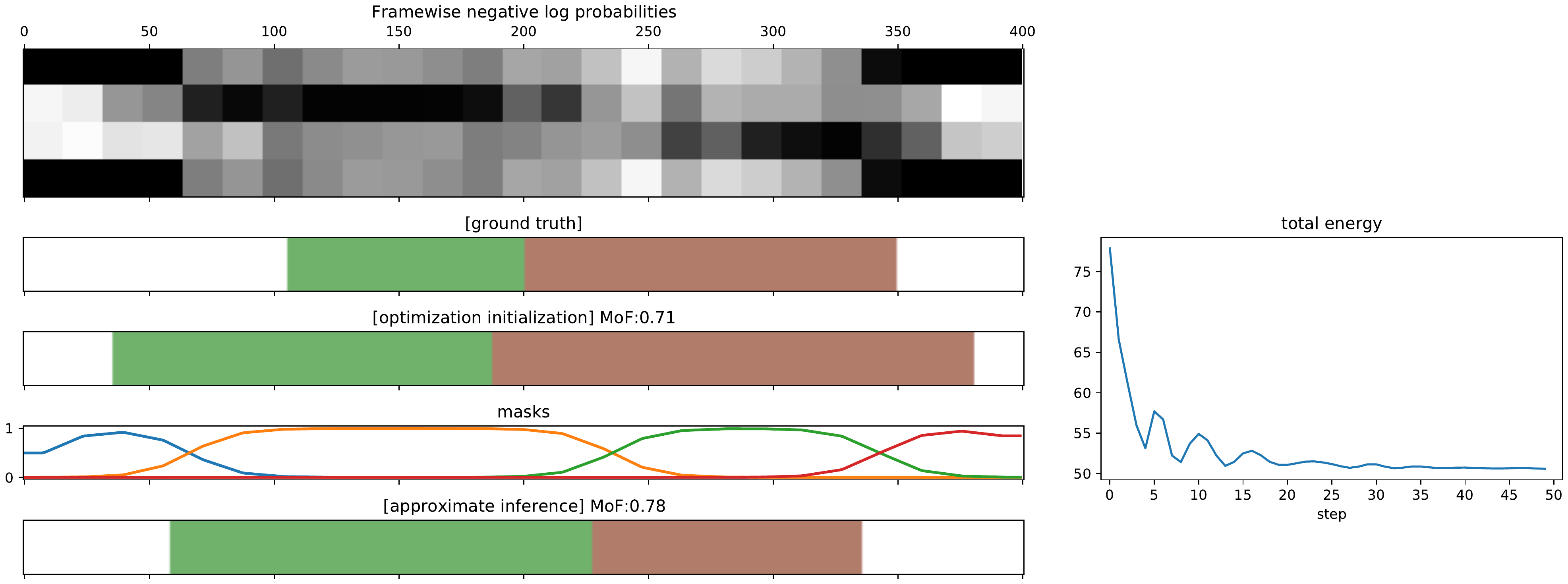}
      \caption{Qualitativ Result: Weakly supervised action segmentation, FIFA + MuCon}
      \label{fig:q:mucon:1}
\end{figure*}

\begin{figure*}
     \centering
      \includegraphics[width=1.0\linewidth]{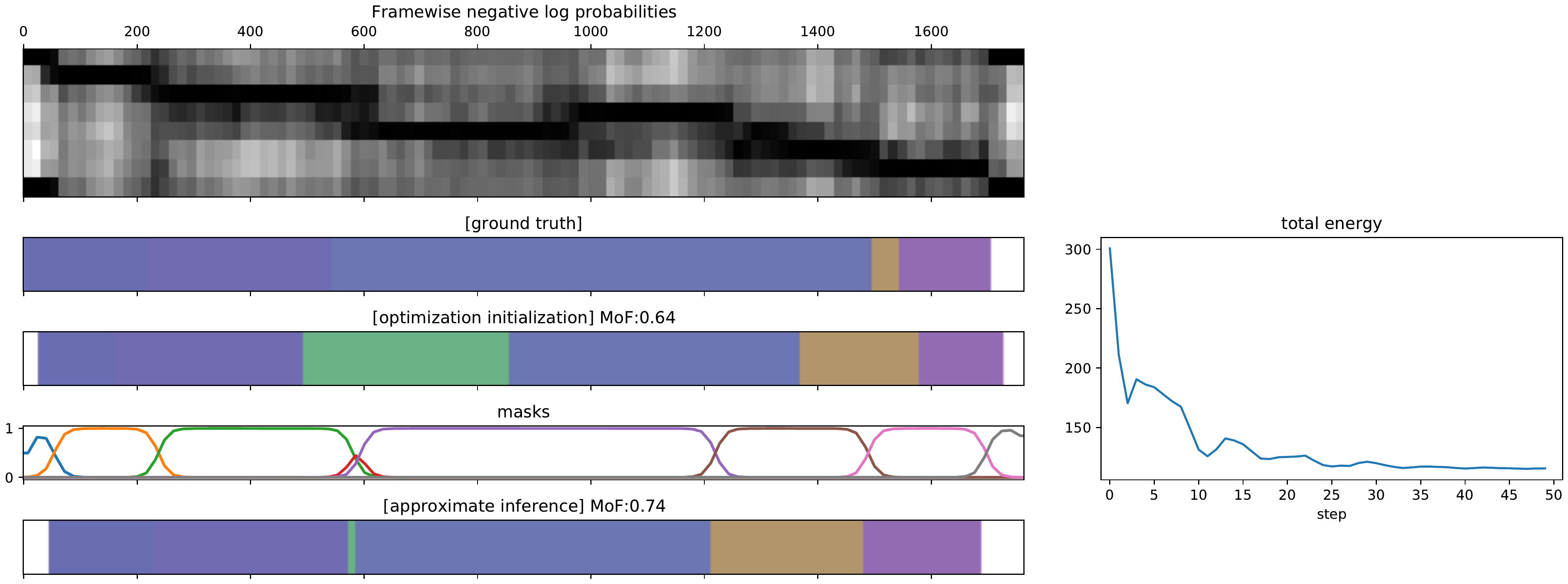}
      \caption{Qualitativ Result: Weakly supervised action segmentation, FIFA + MuCon}
      \label{fig:q:mucon:2}
\end{figure*}

\begin{figure*}
     \centering
      \includegraphics[width=1.0\linewidth]{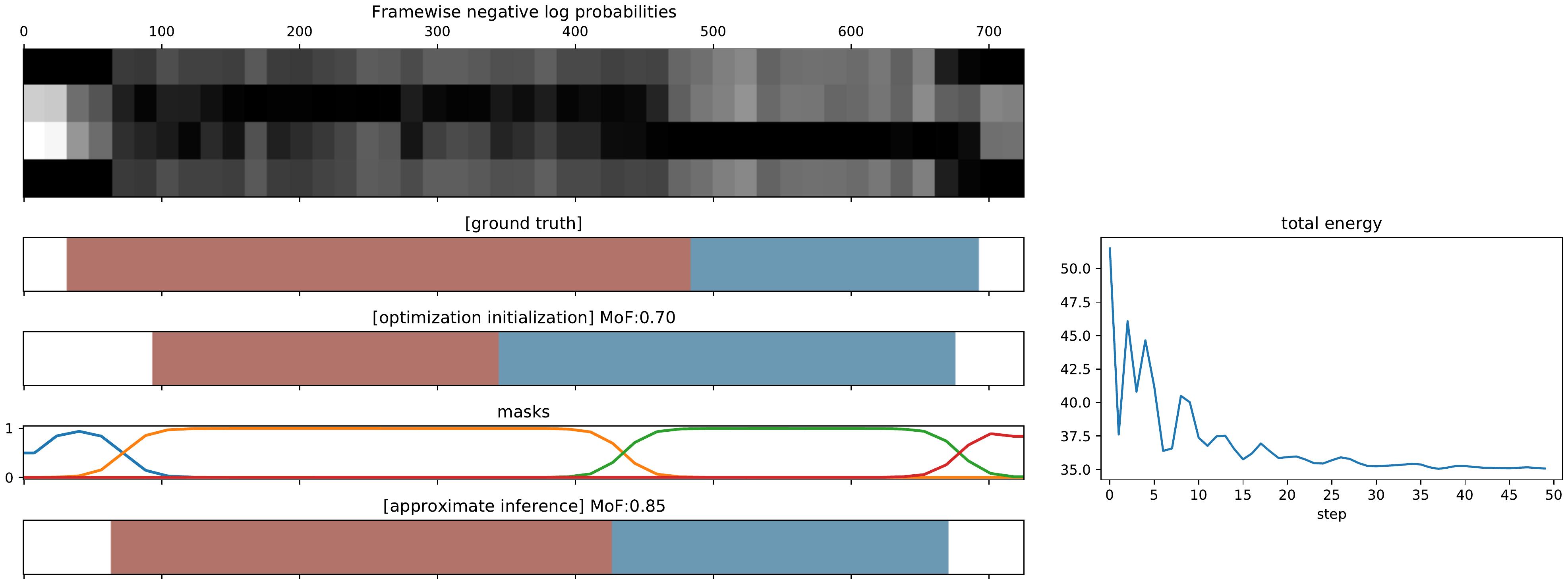}
      \caption{Qualitativ Result: Weakly supervised action segmentation, FIFA + MuCon}
      \label{fig:q:mucon:3}
\end{figure*}

\begin{figure*}
     \centering
      \includegraphics[width=1.0\linewidth]{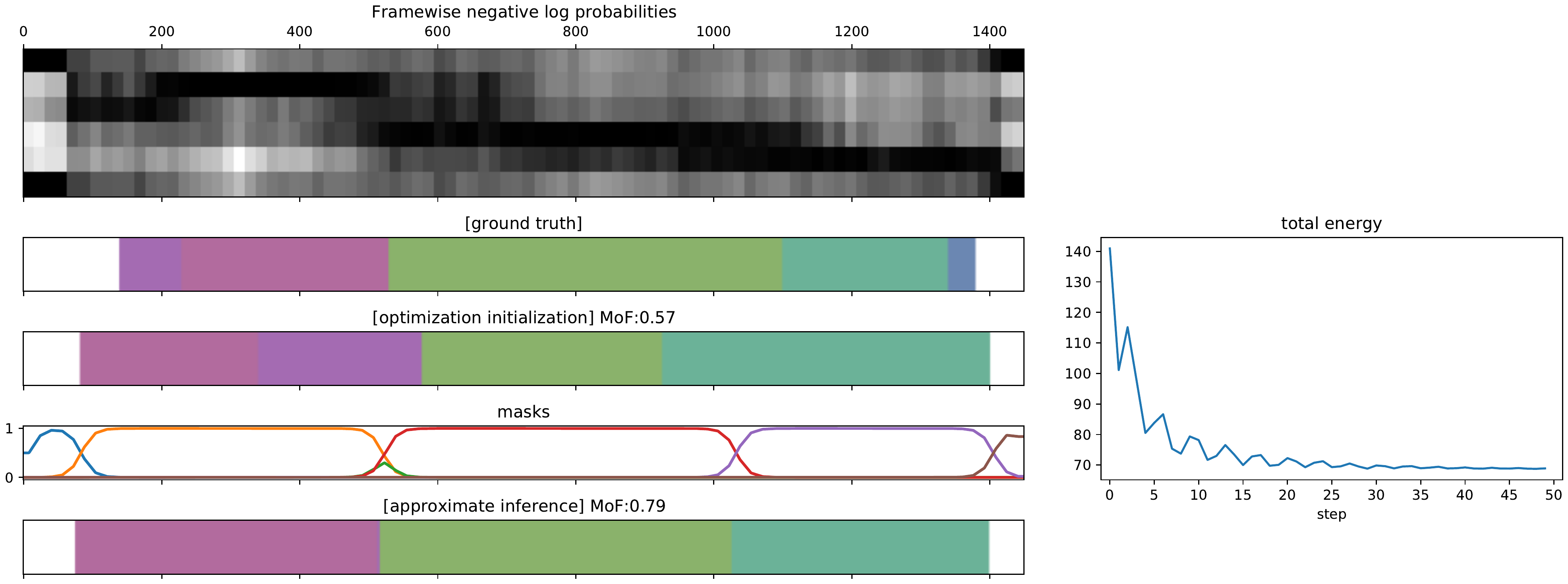}
      \caption{Qualitativ Result: Weakly supervised action segmentation, FIFA + MuCon}
      \label{fig:q:mucon:4}
\end{figure*}

\begin{figure*}
     \centering
      \includegraphics[width=1.0\linewidth]{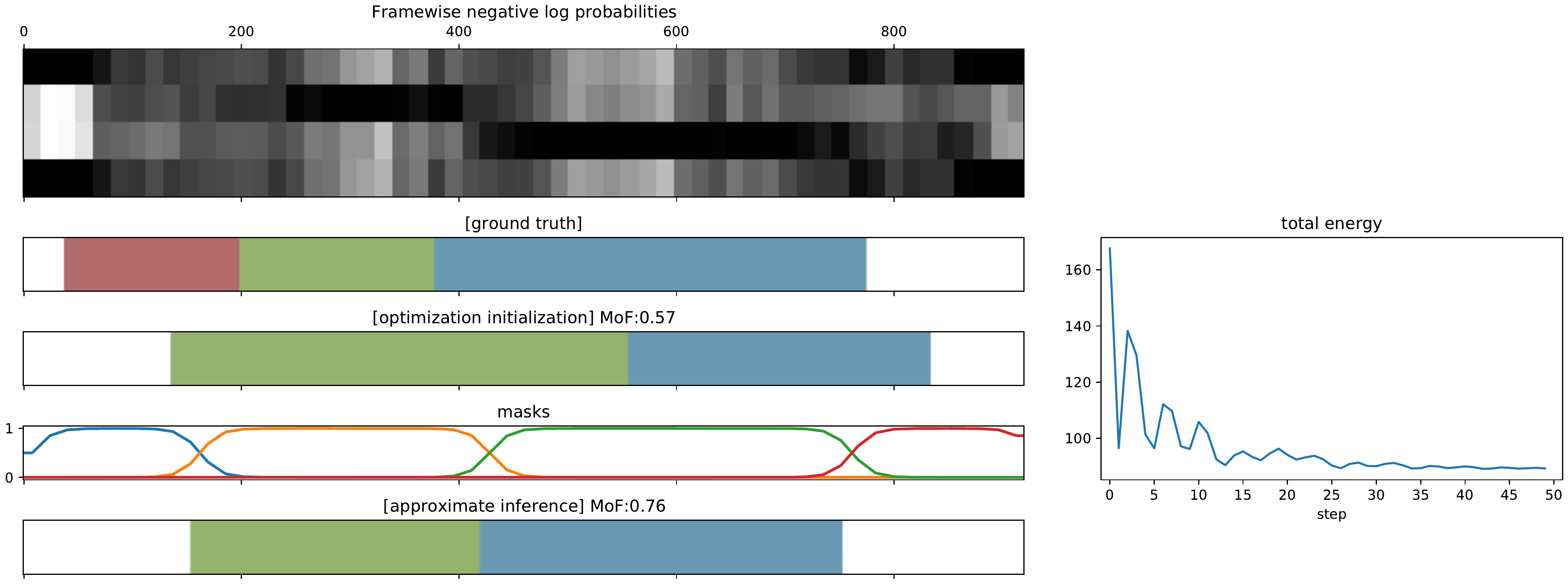}
      \caption{Qualitativ Result: Weakly supervised action segmentation, FIFA + MuCon}
      \label{fig:q:mucon:5}
\end{figure*}

\begin{figure*}
     \centering
      \includegraphics[width=1.0\linewidth]{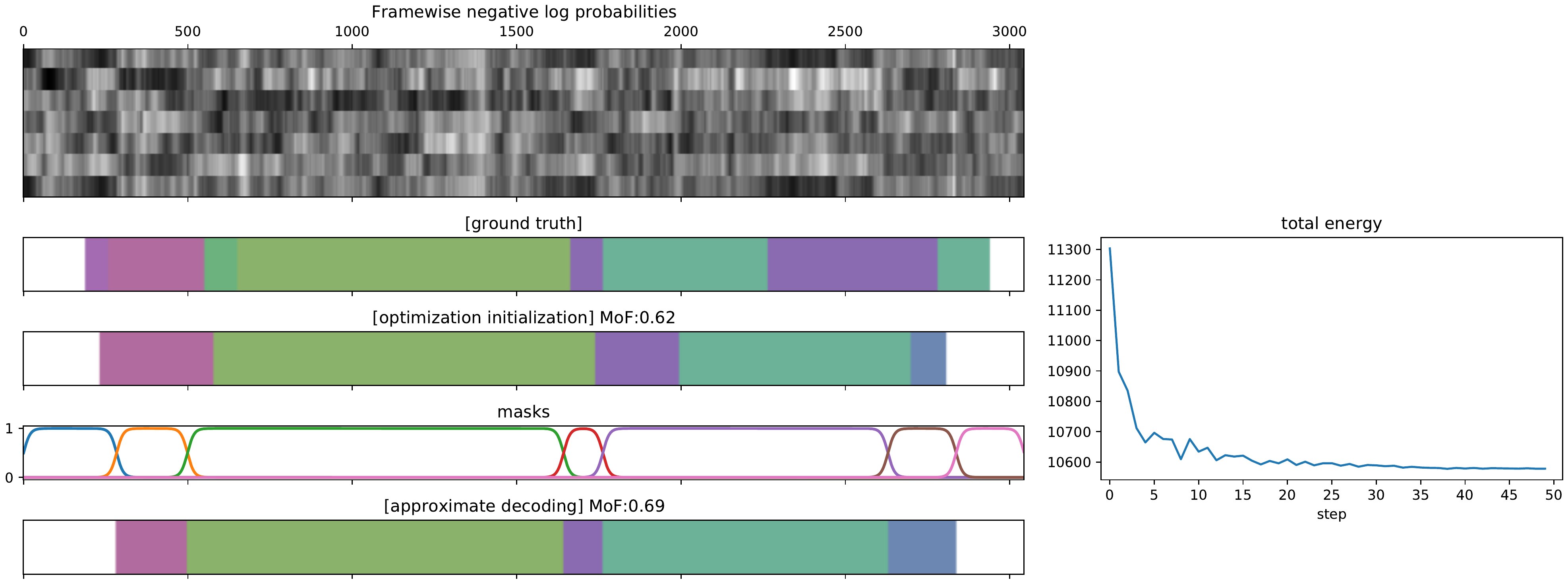}
      \caption{Qualitativ Result: Weakly supervised action segmentation, FIFA + CDFL}
      \label{fig:q:cdfl:1}
\end{figure*}

\begin{figure*}
     \centering
      \includegraphics[width=1.0\linewidth]{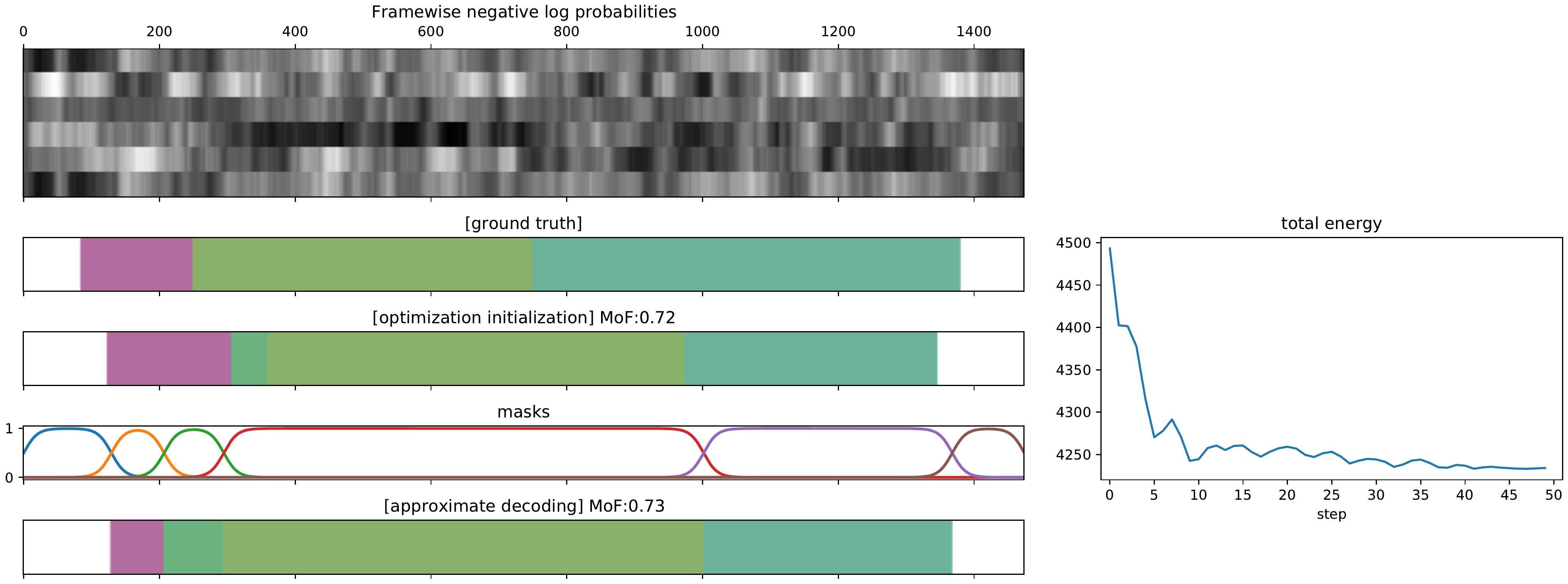}
      \caption{Qualitativ Result: Weakly supervised action segmentation, FIFA + CDFL}
      \label{fig:q:cdfl:2}
\end{figure*}

\begin{figure*}
     \centering
      \includegraphics[width=1.0\linewidth]{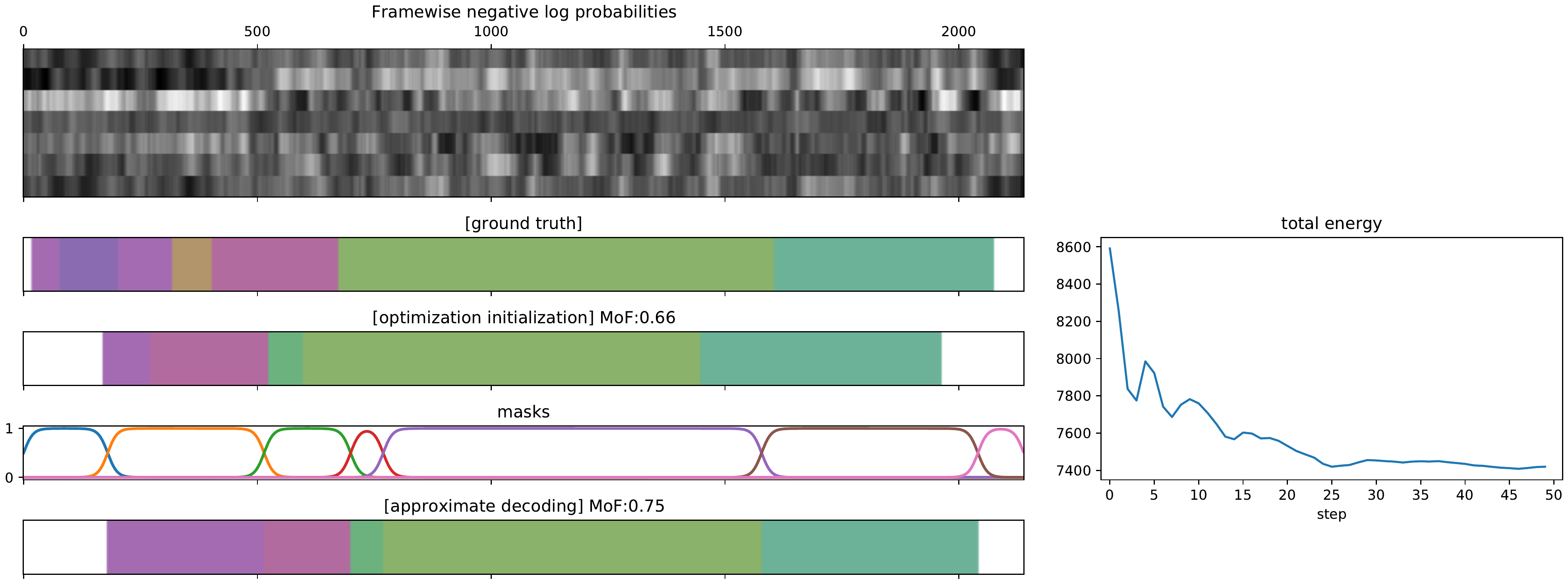}
      \caption{Qualitativ Result: Weakly supervised action segmentation, FIFA + CDFL}
      \label{fig:q:cdfl:3}
\end{figure*}

\begin{figure*}
     \centering
      \includegraphics[width=1.0\linewidth]{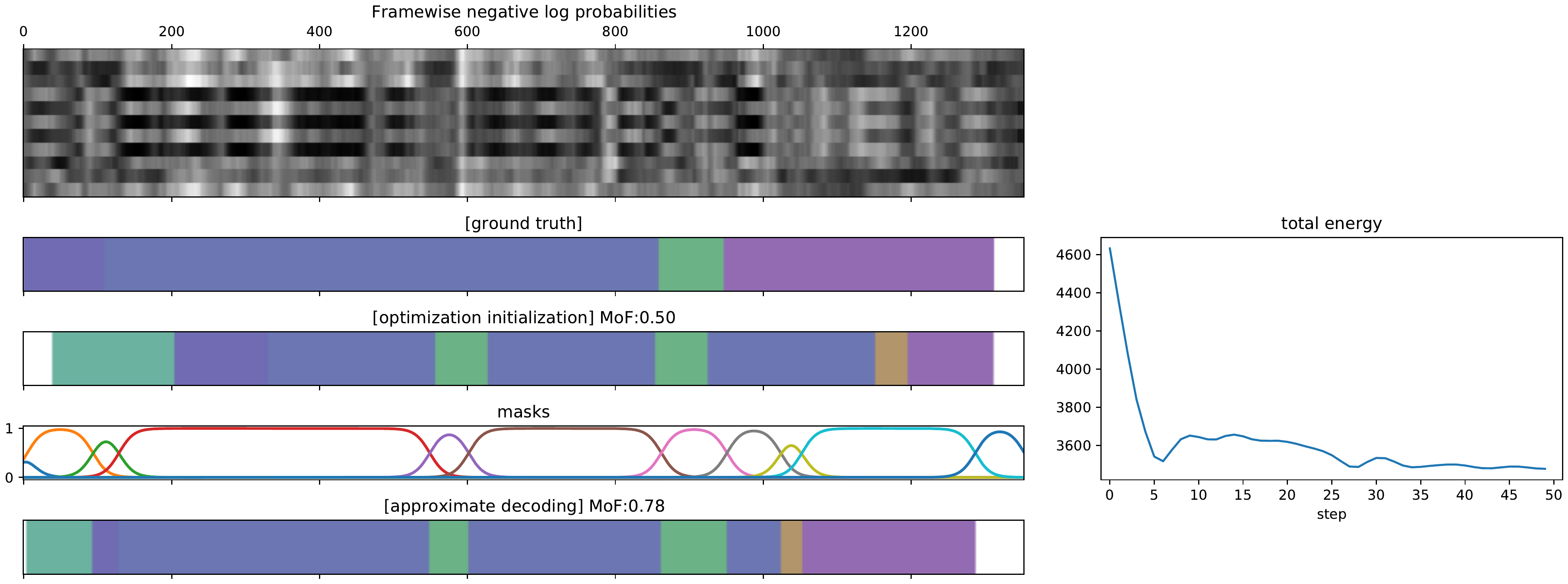}
      \caption{Qualitativ Result: Weakly supervised action segmentation, FIFA + CDFL}
      \label{fig:q:cdfl:4}
\end{figure*}

\begin{figure*}
     \centering
      \includegraphics[width=1.0\linewidth]{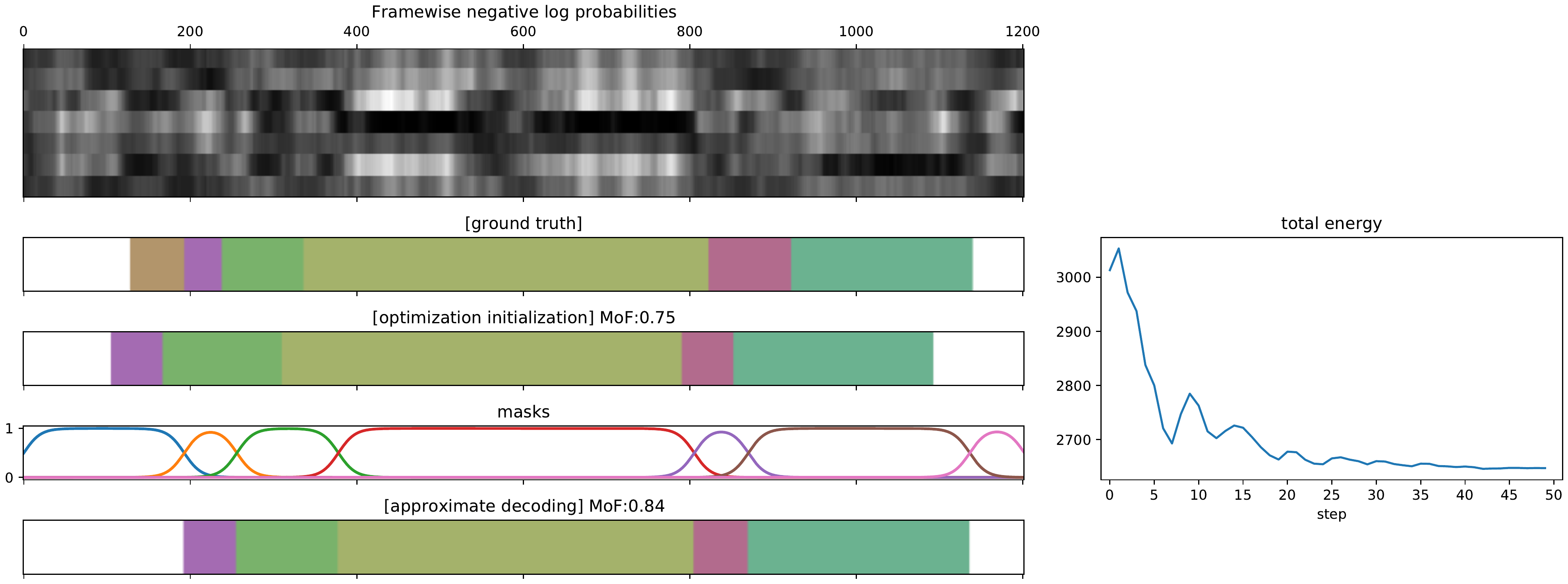}
      \caption{Qualitativ Result: Weakly supervised action segmentation, FIFA + CDFL}
      \label{fig:q:cdfl:5}
\end{figure*}

\begin{figure*}
     \centering
      \includegraphics[width=1.0\linewidth]{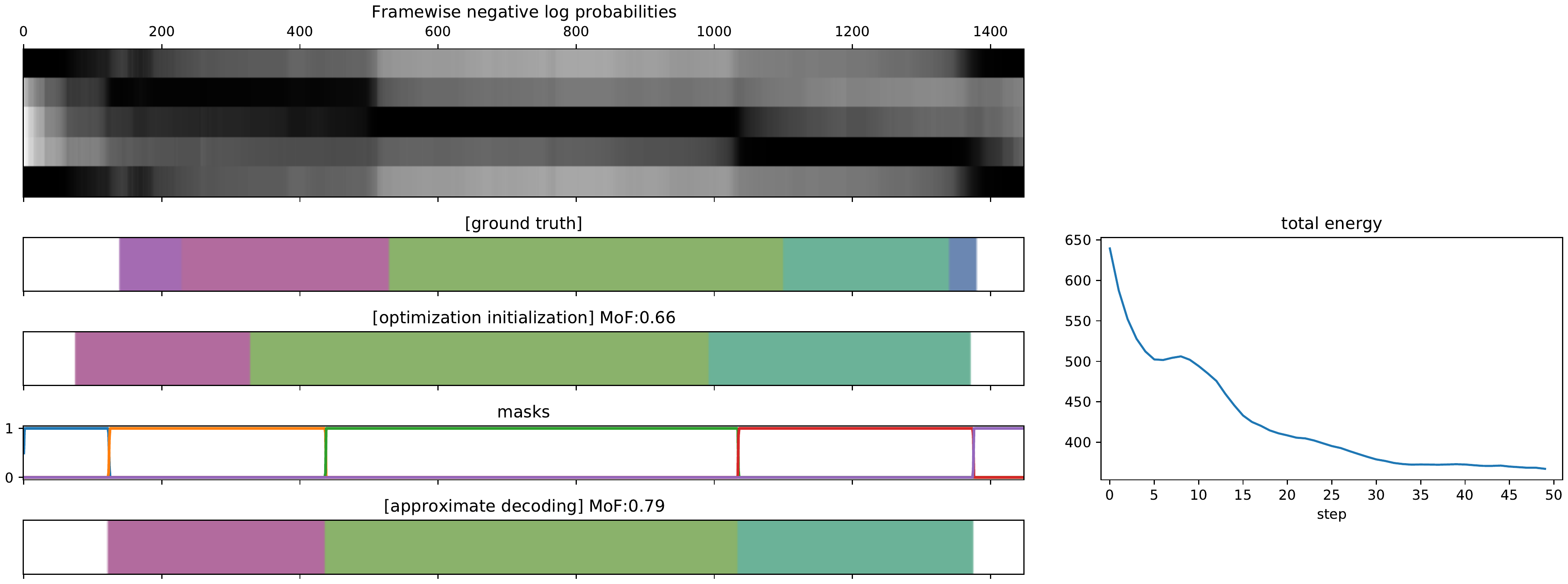}
      \caption{Qualitativ Result: Fully supervised action segmentation, FIFA + MSTCN}
      \label{fig:q:mstcn:1}
\end{figure*}

\begin{figure*}
     \centering
      \includegraphics[width=1.0\linewidth]{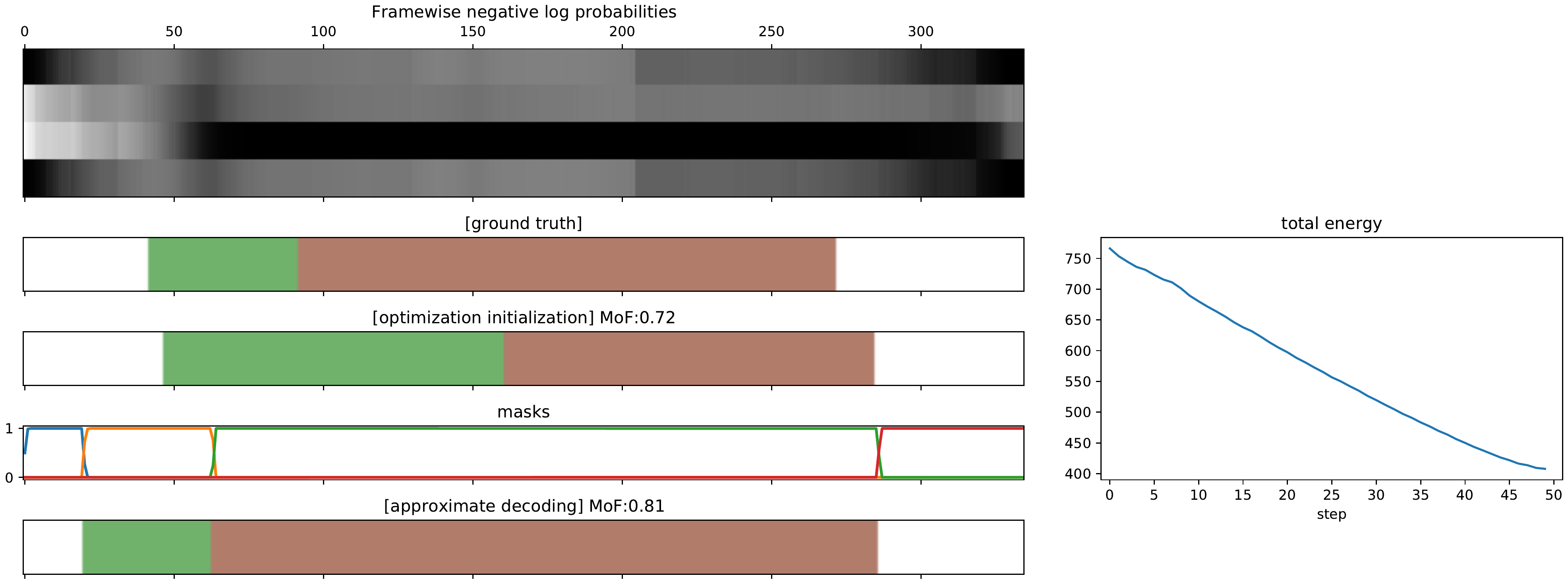}
      \caption{Qualitativ Result: Fully supervised action segmentation, FIFA + MSTCN}
      \label{fig:q:mstcn:2}
\end{figure*}

\begin{figure*}
     \centering
      \includegraphics[width=1.0\linewidth]{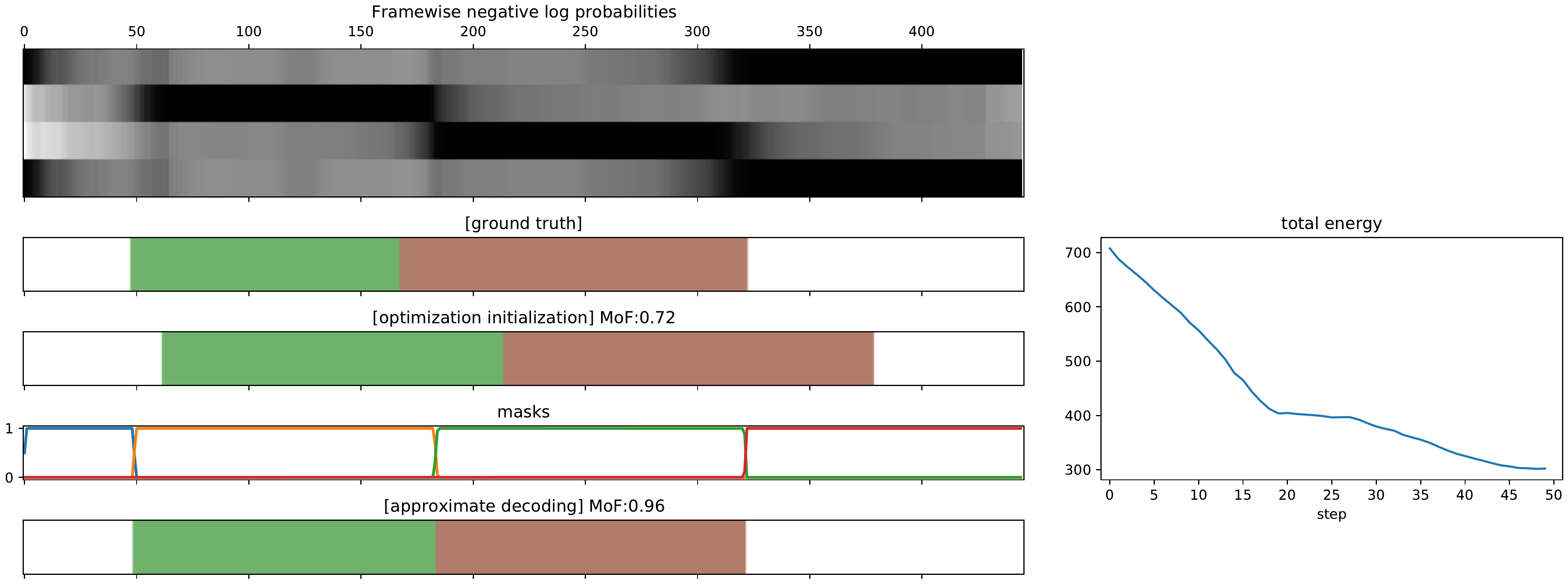}
      \caption{Qualitativ Result: Fully supervised action segmentation, FIFA + MSTCN}
      \label{fig:q:mstcn:3}
\end{figure*}

\begin{figure*}
     \centering
      \includegraphics[width=1.0\linewidth]{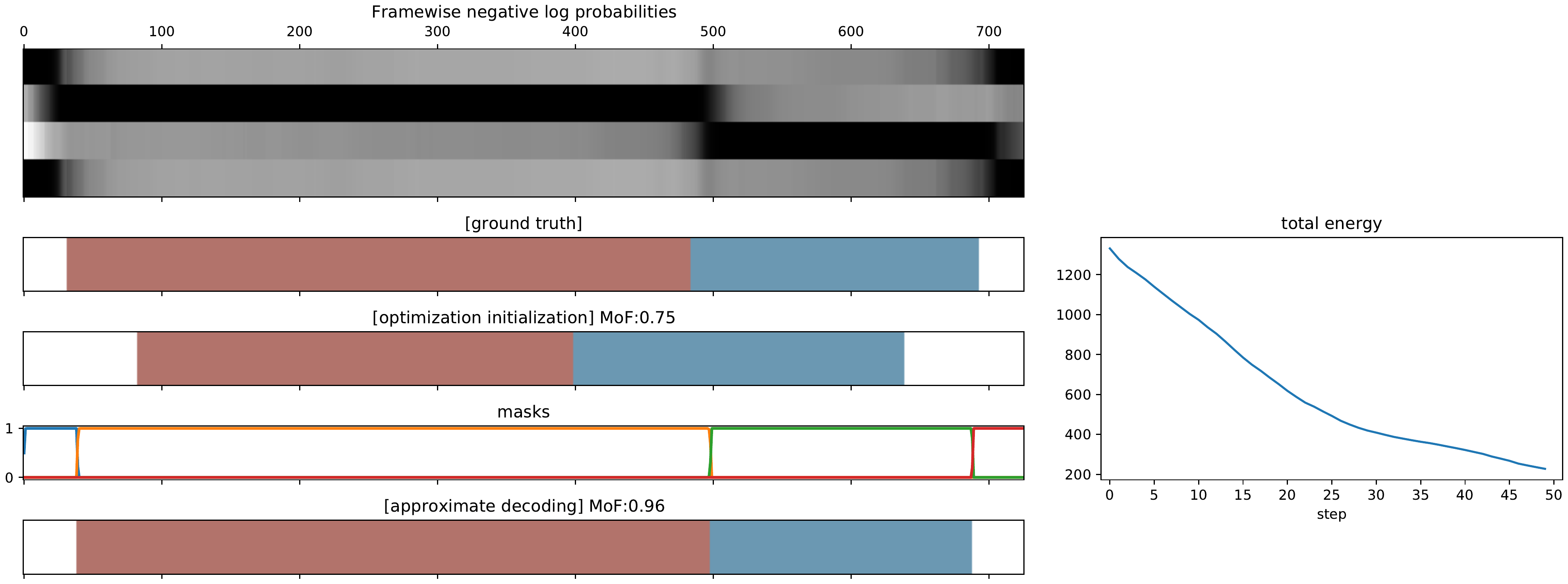}
      \caption{Qualitativ Result: Fully supervised action segmentation, FIFA + MSTCN}
      \label{fig:q:mstcn:4}
\end{figure*}

\begin{figure*}
     \centering
      \includegraphics[width=1.0\linewidth]{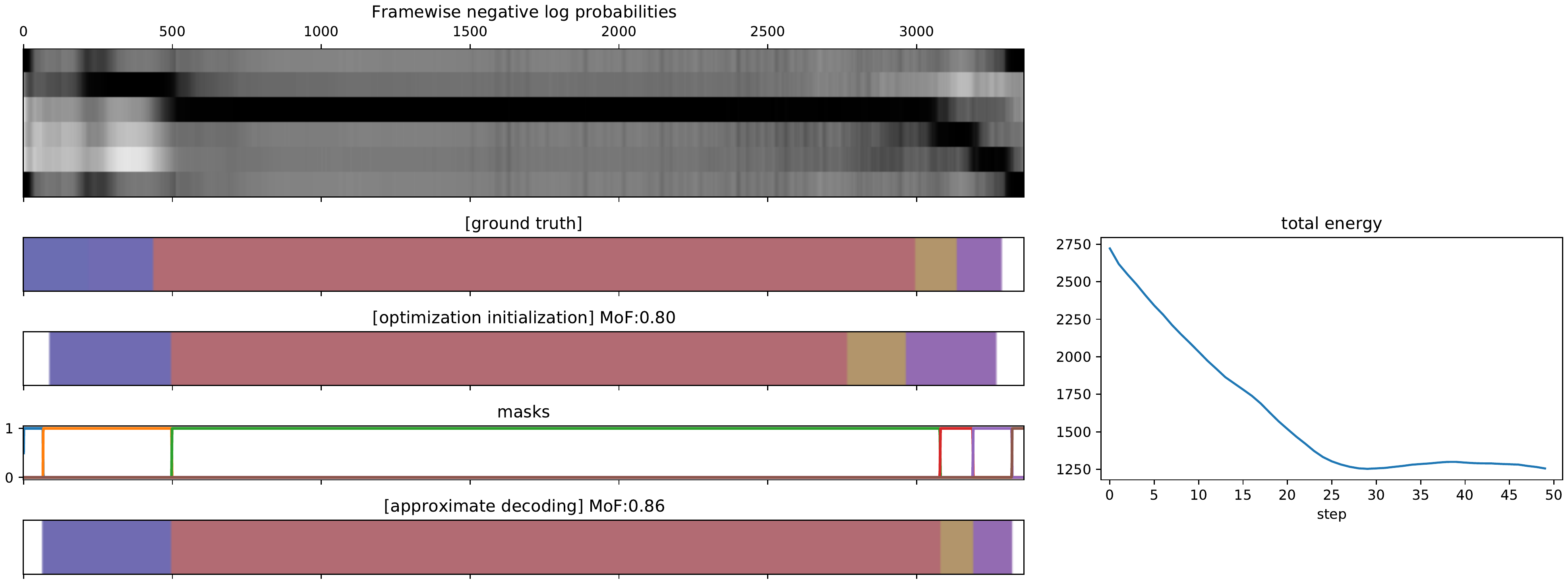}
      \caption{Qualitativ Result: Fully supervised action segmentation, FIFA + MSTCN}
      \label{fig:q:mstcn:5}
\end{figure*}

Figures \ref{fig:q:mucon:1}-\ref{fig:q:mucon:5} show qualitative examples of applying FIFA on top of MuCon \cite{mucon} for weakly supervised action segmentation.
Figures \ref{fig:q:cdfl:1}-\ref{fig:q:cdfl:5} show qualitative examples of applying FIFA on top of CDFL \cite{CDFL} for weakly supervised action segmentation.
Figures \ref{fig:q:mstcn:1}-\ref{fig:q:mstcn:5} show qualitative examples of applying FIFA on top of MSTCN \cite{MS-TCN} for fully supervised action segmentation.

\subsection{Failure Cases}
\begin{figure*}
     \centering
      \includegraphics[width=1.0\linewidth]{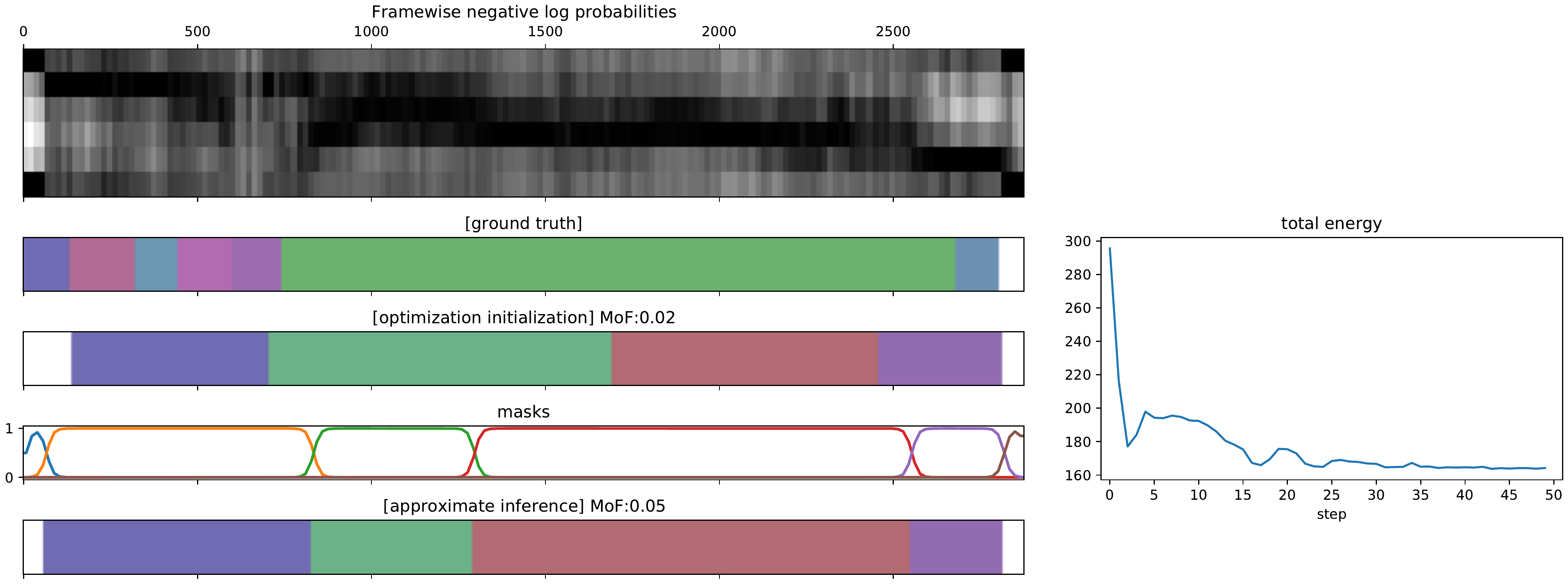}
      \caption{Qualitativ Result: Weakly supervised action segmentation, FIFA + MuCon, Failure Case}
      \label{fig:q:mucon:1:fail}
\end{figure*}

\begin{figure*}
     \centering
      \includegraphics[width=1.0\linewidth]{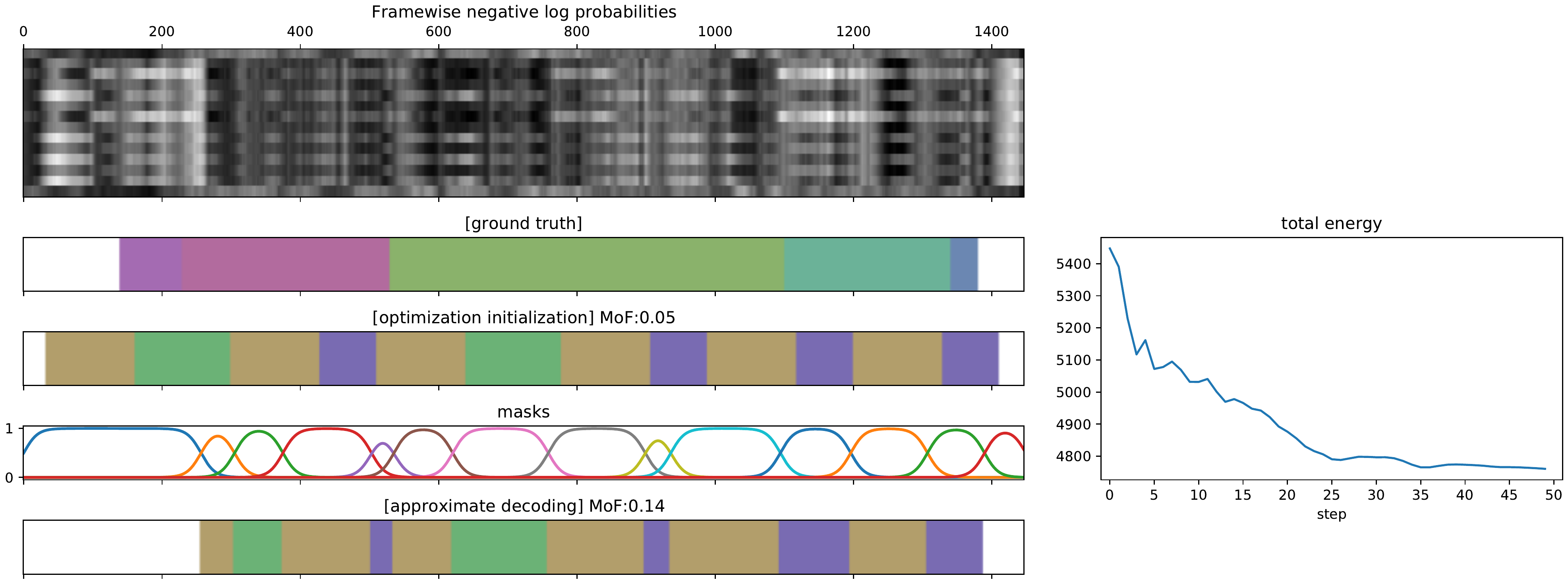}
      \caption{Qualitativ Result: Weakly supervised action segmentation, FIFA + CDFL, Failure Case}
      \label{fig:q:cdfl:1:fail}
\end{figure*}

\begin{figure*}
     \centering
      \includegraphics[width=1.0\linewidth]{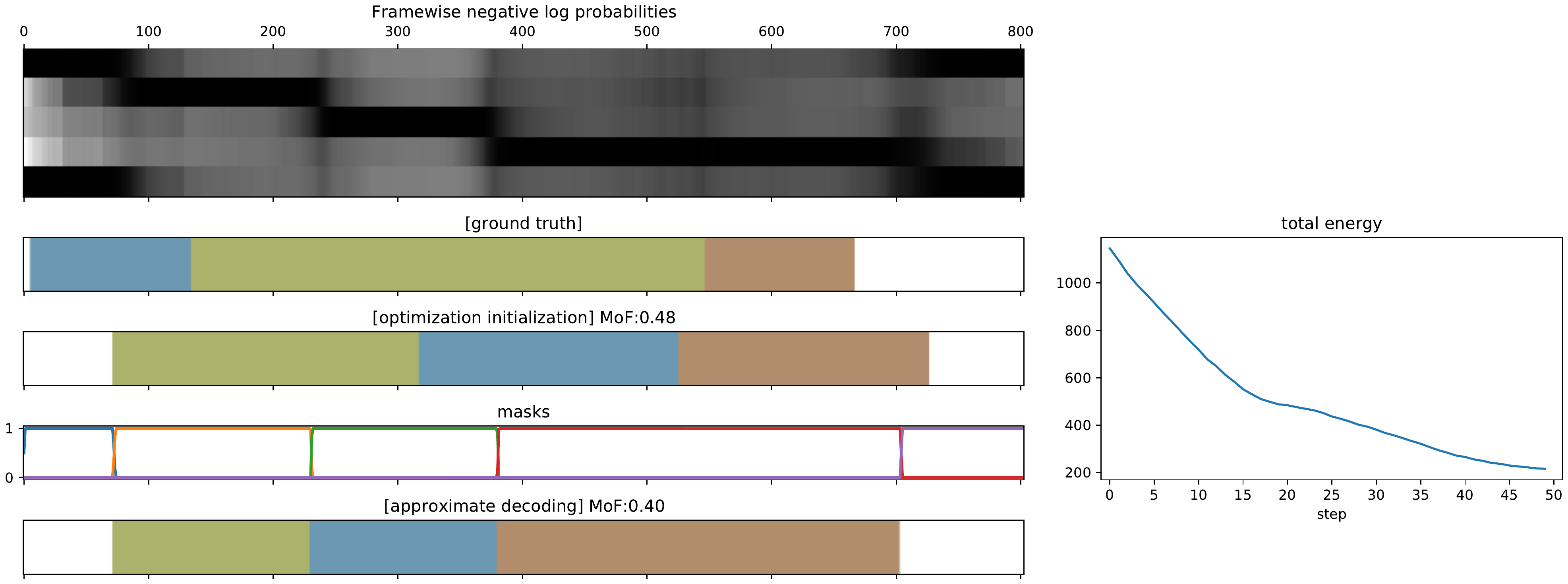}
      \caption{Qualitativ Result: Fully supervised action segmentation, FIFA + MSTCN, Failure Case}
      \label{fig:q:mstcn:1:fail}
\end{figure*}

Figures \ref{fig:q:mucon:1:fail}, \ref{fig:q:cdfl:1:fail} and, \ref{fig:q:mstcn:1:fail} show failure cases for FIFA + MuCon, FIFA + CDFL and, FIFA + MSTCN respectively.
We observe that the major failure case is when the optimization is initialized with an incorrect transcript (Figures \ref{fig:q:mucon:1:fail} and \ref{fig:q:cdfl:1:fail}).
Another failure mode is when the predicted negative log probabilities are not correct (Figure \ref{fig:q:mstcn:1:fail}) which causes the boundaries of actions to be in the wrong location.




\end{document}